\documentclass[11pt]{article}
\usepackage{graphicx}
\usepackage{subfigure}
\usepackage{authblk}
\usepackage{indentfirst,csquotes}

\usepackage{amssymb,amsthm,amsmath}
\usepackage{xcolor,paralist,hyperref,titlesec,fancyhdr,etoolbox}

\usepackage{amsmath}
\usepackage{amsthm}
\usepackage{mathrsfs}
\usepackage{amsfonts}
\usepackage{adjustbox}
\usepackage{algorithmic}
\usepackage[ruled,vlined]{algorithm2e}
\usepackage{array,makecell,multirow}
\hypersetup{ colorlinks=true, linkcolor=black, filecolor=black, urlcolor=black }

\title{Value Function Initialization for Knowledge Transfer and Jump-start in Deep Reinforcement Learning} 
\author{Soumia Mehimeh}
\affil{Non affiliated}
\date{}
\begin{document}

\maketitle

\begin{abstract}
Value function initialization (VFI) is an effective way to achieve a jumpstart in reinforcement learning (RL) by leveraging value estimates from prior tasks. While this approach is well established in tabular settings, extending it to deep reinforcement learning (DRL) poses challenges due to the continuous nature of the state–action space, the noisy approximations of neural networks, and the impracticality of storing all past models for reuse. In this work, we address these challenges and introduce DQInit, a method that adapts value function initialization to DRL. DQInit reuses compact tabular Q-values extracted from previously solved tasks as a transferable knowledge base. It employs a knownness-based mechanism to softly integrate these transferred values into underexplored regions and gradually shift toward the agent’s learned estimates, avoiding the limitations of fixed time decay. Our approach offers a novel perspective on knowledge transfer in DRL by relying solely on value estimates rather than policies or demonstrations, effectively combining the strengths of jumpstart RL and policy distillation while mitigating their drawbacks. Experiments across multiple continuous control tasks demonstrate that DQInit consistently improves early learning efficiency, stability, and overall performance compared to standard initialization and existing transfer techniques.
\end{abstract}
\section{Introduction}
Value function initialization is a simple yet effective transfer learning strategy in reinforcement learning (RL), known for improving jumpstart performance when learning new tasks \cite{maxqinit, mehimeh2023value, mehimeh2025}. This approach is particularly valuable in lifelong learning, where an agent must continually adapt to a distribution of related tasks while reusing knowledge from previous ones. While such strategies have been studied in tabular settings—where value estimates are stored and reused explicitly based on state-action visitation—they remain largely unexplored in DRL.

In this work, we aim to bring the benefits of value VFI, as demonstrated in tabular settings, to DRL. We focus particularly on lifelong learning or any other scenarios involving multiple prior tasks. We consider an agent learning over a distribution of tasks in continuous domains, where each task differs in its underlying dynamics. For each novel task, the agent reuses knowledge extracted from previously solved tasks to initialize the value function and improve early learning efficiency. However, initializing value functions in DRL presents several challenges.

The first challenge is the infeasibility of explicitly storing values for all individual state-action pairs in continuous spaces. A simple solution would involve saving the full value networks of all previously learned tasks and querying them to extract value estimates for new inputs, as seen in prior work on distillation \cite{policydistillation}. However, this approach contradicts the goals of lifelong learning, where the agent is expected to retain a compact and scalable knowledge base rather than accumulate a growing archive of neural models \cite{khetarpal2022towards}.

The second challenge concerns the instability of value approximations across tasks. Neural networks in DRL act as function approximators, but they are highly sensitive to even small changes in task dynamics. Slight differences in transition probabilities or reward structures can result in substantially different Q-value outputs for the same states, even when the optimal policies remain functionally similar \cite{bellemare2019geometric}. These inconsistencies are not due to policy shifts, but rather to the optimization trajectory and generalization behavior of deep networks \cite{wang2024investigating}. As a result, transferred Q-values from prior networks may be either overestimated or underestimated in unpredictable ways, making them unreliable for reuse in new tasks.

The third challenge lies in both localizing and controlling the transferred knowledge during training. In tabular RL, each state-action pair is explicitly initialized and updated as it is visited, which ensures that all parts of the state-action space benefit from initialization. In contrast, neural networks generalize across the input space, making it difficult to ensure that each state-action pair receives its share of transferred knowledge, especially in continuous domains, where exact repetitions are rare. At the same time, we must also control the influence of the transferred values so they guide only the early phase of learning. Without a proper decay mechanism, initialization can unintentionally persist throughout training, interfering with the agent’s ability to adapt. Prior work, such as Jumpstart Reinforcement Learning (JSRL) \cite{uchendu2023jump}, addresses this by applying time-based decay to reduce the influence of an expert policy. While this helps limit the duration of guidance, it introduces a bias: states encountered early benefit more, while important states visited later may receive little to no guidance.

To overcome these limitations, we propose \textbf{D}eep \textbf{Q}-values \textbf{Init}ialization ($\textsc{DQInit}$), a VFI method for DRL that enables effective transfer of prior knowledge in a scalable and stable way. Instead of relying on stored models or policy distillation, $\textsc{DQInit}$ uses a compact tabular knowledge base containing value estimates from previously solved tasks, discretized or clustered over the continuous input space. To integrate these values into the agent’s learning process, we introduce an adaptive initialization function based on the notion of knownness—a measure of how frequently each state-action pair has been visited in the current task. This function softly integrates transferred Q-values in underexplored regions and gradually shifts toward using the agent’s learned estimates as experience accumulates. In this way, we avoid hard-coded time decay and instead guide learning based on what the agent has and hasn’t yet seen.

$\textsc{DQInit}$ can be used in three modes: (1) soft policy guidance using the initialized value function, (2) as an auxiliary loss to regularize the value function, and (3) as a distillation to guide policy learning. Positioned between policy guidance methods \cite{uchendu2023jump} and policy distillation \cite{policydistillation}, our method avoids their limitations by using transferred Q-values only for initialization, allowing the agent to freely learn from the environment without being constrained by full imitation or ongoing advice. Our experiments confirm that $\textsc{DQInit}$ consistently improves jumpstart performance and stability compared to standard initialization, JSRL, and policy distillation baselines. Moreover, it provides empirical validation for theoretical insights from tabular VFI, showing that their core principles can be successfully extended to deep RL in continuous domains.
\section{Related Works}
\subsubsection*{Value function initialization} VFI has previously been proposed as a means to accelerate learning in novel tasks within lifelong reinforcement learning, by leveraging distributional information from prior tasks to estimate a strong prior for the value function. One such approach, MaxQInit\cite{maxqinit}, initializes the value function optimistically using the maximum observed value, which is justified by its ability to preserve polynomial optimality guarantees under the MDP framework. However, this method assumes a uniform task distribution, which was later shown by \cite{mehimeh2023value} to be problematic in the presence of outlier tasks, where optimistic initialization can degrade performance. To address this, they proposed UCOI as an improvement over MaxQInit by introducing a trade-off between optimistic initialization and the empirical expectation of past data. This trade-off is modulated by confidence estimates from the PAC-MDP framework and uncertainty derived from the task distribution. The approach has demonstrated improved performance in scenarios involving non-uniform goal distributions.

In a related direction, \cite{mehimeh2025} proposed initializing the value function based on log-normality assumptions. Their approach is motivated by the observation that, under sparse rewards and normally distributed tasks, the resulting Q-values tend to follow a log-normal distribution. This method was shown to yield more accurate value estimates than both UCOI and MaxQInit. 

However, the aforementioned VFI methods have so far been evaluated only in tabular settings, and their applicability to deep reinforcement learning remains unexamined. In this work, we aim to implement and evaluate these VFI strategies in DRL to investigate their generalizability and effectiveness beyond the tabular setting.

\subsubsection*{Policy guidance} Our work is related to policy guidance methods, which aim to facilitate efficient learning, particularly during the early stages of training. This category includes action advice methods, where agents receive expert suggestions either under a budget \cite{actionadvice} or through a time-decaying mixture of the expert and the learner’s current policy \cite{uchendu2023jump}. It also encompasses imitation learning, which leverages expert demonstrations \cite{zare2024survey}. However, these approaches typically rely on explicit policy transfer or access to demonstrations. In contrast, our method is based on the transfer of value functions, which guide action selection indirectly. Moreover, we propose a mechanism that leverages knowledge inferred from the current task to guide the agent’s reliance on transferred value estimates, thereby eliminating the need for advice budgeting and temporal decay heuristics.

\subsubsection*{Policy distillation} A complementary line of work involves policy distillation, in which the learning objective is regularized by a KL divergence term to mimic expert behavior. Numerous distillation strategies have been developed, including single-expert distillation \cite{policydistillation}, multi-expert aggregation\cite{czarnecki2019distilling}, and dual-policy learning \cite{lai2021dual}. Nevertheless, such approaches typically require either a closely related expert or access to multiple teacher models, making them computationally and memory-intensive. In our work, we address this limitation by replacing the need for multiple expert models with a compact knowledge base, in the form of Q-tables, which enables efficient value-based transfer without incurring the overhead associated with querying multiple policies.

\section{Preliminaries}

We consider a standard Markov Decision Process (MDP), defined by the tuple $(\mathcal{S}, \mathcal{A}, \mathcal{P}, r, \gamma)$, where $\mathcal{S}$ is the set of states, $\mathcal{A}$ the set of actions, $\mathcal{P}(s'|s,a)$ the transition probability, $r(s,a)$ the reward function, and $\gamma \in [0,1)$ the discount factor. The agent's objective is to learn a policy $\pi(s)$ that maximizes the expected discounted return and its value function $Q(s,a)$.

\subsection*{Deep Q-network}: We focus on the Deep Q-Network (DQN) algorithm, which approximates the optimal action-value function $Q^*(s,a)$ using a neural network. The network is trained by minimizing the temporal difference error:
$$
\mathcal{L}_{TD} = \mathbb{E}_{(s,a,r,s')}\left[\left(r + \gamma \max_{a'} Q(s', a'; \theta^-) - Q(s, a; \theta)\right)^2\right]
$$
Where $\theta^-$ are the parameters of a target network periodically updated from $\theta$.

\subsection*{State-Action Discretization}: To handle continuous state and action spaces in DRL, we define a discretization function
$$
\phi: S \times A \rightarrow \overline{S} \times \overline{A}
$$
where $S \subset \mathbb{R}^n$ and $A \subset \mathbb{R}^m$ are the original continuous spaces, and $\overline{S}, \overline{A}$ are their discretized representations. This mapping enables compact storage and reuse of value information by abstracting the continuous domain into a finite set of bins.

\subsection*{Problem setting:} We consider a lifelong reinforcement learning setting, where the agent encounters a sequence of tasks sampled from a distribution $\Omega$, with each task $M_i \sim \Omega$ differing in reward or dynamics. The agent has previously solved $n$ tasks and stored their value functions in a set $\mathcal{Q} = \{Q_i(s,a)\}_{i=1}^n$. Given an initialization method $\mathscr{A}_{init}$, the agent derives an initial value function $Q^{\emptyset}$ for a new task. This function is used to transfer knowledge to future tasks from the same distribution.
\section{Methodology}

The $\textsc{DQInit}$ objective is to use $Q^{\emptyset}(s, a)$ to guide the agent’s early-stage learning. This guidance must satisfy two conditions: (1) it should primarily influence the initial phase of learning the policy or the value $Q^{\theta}$ of the state-action and gradually phases out, and (2) it should be applied to every state-action pair, regardless of when it is first encountered during training.

To address the first condition, a natural approach is to blend the transferred and learned values using a convex combination, for example: 
$\widetilde{Q}(s, a) = (1 - f) Q^{\emptyset}(s, a) + f Q^{\theta}(s, a),$ where $Q^{\theta}$ is the agent’s learned value function and $f \in [0, 1]$ controls the balance. However, using a \textit{time-based decay} for $f$—for instance, based on training steps or episodes—fails to satisfy the second condition: states encountered early receive strong guidance, while those visited later may receive none, regardless of their novelty or importance \cite{uchendu2023jump}.

To overcome this, we replace the time-based decay with a task-specific function that quantifies how well the agent \textit{knows} a state-action pair by tracking its visitation. We define the notion of “knownness”—a quantity that starts at 0 for unvisited pairs and increases toward 1 as a pair is encountered more frequently and becomes familiar to the agent . Since explicit counting over continuous state-action spaces is infeasible, we adopt a similarity-based grouping mechanism that clusters similar state-action pairs. The latter method is commonly used in count-based exploration for DRL, aligns naturally with our goal of modulating the influence of initialization based on familiarity rather than time. In the following we introduce all the elements that are part of our algorithm, starting with the adaptive function, the source of knowledge that produces Q empty state and the knownness function.
\subsection{The adaptive function} 
the adaptivity mechanism of our approach DQInit is highly based on an "adaptive function" $\widetilde Q$ that operates as follows:

\begin{equation}
\widetilde{Q}(s,a) = K(s,a)\, Q^{\theta}(s,a) + \left(1 - K(s,a)\right)\, Q^{\emptyset}(s,a)
\end{equation}

Where:
\begin{itemize}
    \item $Q^{\theta}(s,a)$ is the Q-function produced by a neural network parameterized by $\theta$,
    \item $Q^{\emptyset}(s,a)$ is an initialization function derived from prior knowledge (e.g., expert value functions),
    \item $K: S \times A \rightarrow [0,1]$ quantifies the knownness of a state-action pair.
\end{itemize}

This formulation ensures that when knownness is low, the agent primarily relies on the initialization function $Q^{\emptyset}(s,a)$, leveraging prior knowledge acquired from previously solved tasks. As more interactions are collected and knownness increases, the agent gradually shifts toward relying on its own learned Q-function $Q^{\theta}(s,a)$, thereby phasing out the influence of the initialization.

\subsection{Initial Value Function}

We consider two sources for constructing the initialization function $Q^{\emptyset}(s,a)$: (1) aggregating predictions from previously trained neural models, and (2) using a tabular value function obtained through discretization.
\subsubsection{From Neural Models}

Each past task yields a trained Q-function $Q_i^{\theta}$. For a given state-action pair $(s,a)$, we collect the corresponding predictions $
\mathcal{Q}^{\theta}(s,a) = \{Q_i^{\theta}(s,a)\}_{i=1}^n$. An initialization function $\mathscr{A}_{\text{init}}$ is then used to compute the initial function as $
Q^{\emptyset}(s,a) = \mathscr{A}_{\text{init}}(\mathcal{Q}^{\theta}(s,a))
$

Why retaining knowledge from models is troublesome? We notice two major limitations of considering querying models to compute the initial function
\begin{itemize}
\item \textit{Model storage overhead:} since state and action are continuous, we must query the previous models to get the value of each model, especially if we consider that the tasks are dissimilar and we don't know which one to outcast and which one to keep retaining knowledge from. however as number of tasks evolve the number of models increases and querying them becomes computationally infeasible and time and storage consuming. in settings such as lifelong learning, the objective is to consider a compact knowledge base and therefore this method is not suitable for this setting. \cite{khetarpal2022towards}.
\item \textit{Sensitivity to dynamics:} Neural networks are highly sensitive to training variations and task dynamics \cite{osband2018randomized,henderson2018deep}. Even minor changes in environment parameters can produce inconsistent Q-values, despite similar policies \cite{bellemare2019geometric,wang2024investigating}.
\end{itemize}
\subsubsection{From Tabular Value Functions}

To address these limitations, we propose using a discretized tabular value function as the source $
Q^{\emptyset}(s,a) = \mathscr{A}_{\text{init}}(\mathcal{Q}^{\#}[\phi(s,a)])
$ where $\phi(s,a)$ maps continuous state-action pairs to a finite set, and $Q^{\#} : \overline{S} \times \overline{A} \rightarrow \mathbb{R}$.

The tabular values are learned in parallel during training on each task using standard Q-learning:

\begin{align*}
 Q^{\#}(\phi(s,a))& \leftarrow Q^{\#}(\phi(s,a)) + \alpha [ r(s,a)\\ 
&+ \gamma \max_{a'} Q^{\#}(\phi(s',a')) - Q^{\#}(\phi(s,a))]
\end{align*}

\noindent\textit{\textbf{Why using tabular Q-values as a source of knowledge?}} We assume this form of source knowledge is more reliable for two reasons:
(1) Tabular Q-learning converges more consistently and is less prone to overestimation errors tied to network instabilities or hyperparameters.
(2) Initialization methods actually do not require storing all value tables from past tasks, we maintain only a few number of tables that contains some statistical properties over them —such as $Q^{\#}_{\text{mean}}$ (the average), $Q^{\#}_{\text{std}}$ (as a standard deviation)or $Q_{max}^{\#}$.
These tables are updated as new tasks are learned, allowing the agent to retain useful knowledge from a small, fixed-size representation. This serves the same purpose as keeping all previous models, but with much lower storage and computational cost.

\subsection{Knownness Function}

There exist various ways to quantify knownness, including using distance-based measures to estimate the novelty of a state-action pair relative to previously visited ones \cite{ladosz2022exploration}. In contrast, our formulation is more closely aligned with the count-based method used for \textsc{Rmax} algorithm \cite{rmax}, where a transition is considered known once it has been visited sufficiently often. Specifically, we track visit counts $N(\phi(s, a))$ for each discretized state-action pair \cite{tang2017exploration}. The knownness function is defined as:

$$
K(s,a) = \left[\frac{N(\phi(s, a))}{m}\right]^p
$$

Where $m$ is the maximum visitation threshold and $p$ is a smoothing exponent controlling the rate of increase. Knownness starts at 0 for novel regions and increases toward 1 as familiarity grows, gradually reducing the influence of the initialization.

Parameter selection for $m$ and $p$ depends on the discretization granularity. We empirically analyze their impact by observing reward trends as a function of the knownness ratio across tasks in Table 2 in the supplementary materials.

\section{$\textsc{DQInit}$ for DQN}
After introducing the components of our adaptive function, we now describe how to incorporate \textsc{DQInit} in DQN for learning from prior tasks. As depicted in Algorithm \ref{alg:dqinit}, we use the classical DQN structure, with the addition of three configuration flags, $use_{\widetilde\pi}, use_{\widetilde{\mathcal{L}}}$ and $use_{\mathcal{L}_{KL}}$ that control whether and how the adaptive function is used.
\begin{algorithm}[t]
\caption{DQN+$\textsc{DQInit}$}
\label{alg:dqinit}
\begin{algorithmic}[1]
\REQUIRE Config flags: $use_{\widetilde\pi}, use_{\widetilde{\mathcal{L}}}, use_{\mathcal{L}_{KL}}$
\STATE Initialize DQN parameters, replay buffer $\mathcal{D}$, counters $N(s,a) \leftarrow 0$
\FOR{each episode}
  \STATE Reset environment, observe $s$
  \FOR{each step}
    \STATE Select random action $a$ if $< \epsilon$;
    \STATE else if $use_{\widetilde\pi}$: $a=\arg\max_a \widetilde{Q}(s,a)$  
    \STATE else $a=\arg\max_a Q^{\theta}(s,a)$
    \STATE Execute $a$, get $(r, s')$, store in $\mathcal{D}$, update $N(s,a)$
    \STATE Sample minibatch $\{(s,a,r,s')\}$ from $\mathcal{D}$

    \STATE Compute TD loss: $\mathcal{L}_{TD}$;
    \STATE \textbf{If} : $\widetilde{\mathcal{L}}= MSE(Q^{\theta},\widetilde Q)$ \textbf{else} $\widetilde{\mathcal{L}}=0$;
    \STATE \textbf{If} $use_{\mathcal{L}_{KL}}$: $\mathcal{L}_{KL} \leftarrow \text{KL}(\pi^{\theta} \|\pi^{\emptyset})$ \textbf{else} $\mathcal{L}_{KL}=0$;

    \STATE Update $\theta$ using total loss: $\mathcal{L}_{TD} + \widetilde\lambda \widetilde{\mathcal{L}} + \lambda_{KL} \mathcal{L}_{KL}$;
    \STATE Periodically update DQN target network
  \ENDFOR
\ENDFOR
\end{algorithmic}
\end{algorithm}

\subsubsection{Soft Policy Guidance} When $use_{\widetilde\pi}$ is enabled, the agent selects actions using:

$$
\widetilde{\pi}(s) = \arg\max_a \widetilde{Q}(s,a)
$$

We refer to this mechanism as \textit{soft policy guidance} because, unlike methods such as action reuse or JSRL—which rely on explicit policies or action advice from an expert—it modulates guidance internally through the value function. As the agent gains experience, the influence of the transferred values naturally decays, and the learned value function converges to $Q^{\theta}$, with the resulting policy converging to $\pi^{\theta}$.
\begin{figure*}[ht!]
\centering 
\begin{tabular}{m{0.1em}*{5}{>{\centering\arraybackslash}m{8.5em}}}

&$\widetilde{\pi}$ & $\widetilde\pi,\widetilde{\mathcal{L}}$ & $\widetilde\pi,\mathcal{L}_{KL}$ & $\widetilde\pi,\mathcal{L}_{KL}, \widetilde{\mathcal{L}}$ 
\\
\rotatebox{90}{Mountain Car} &\includegraphics[width=1 \linewidth]{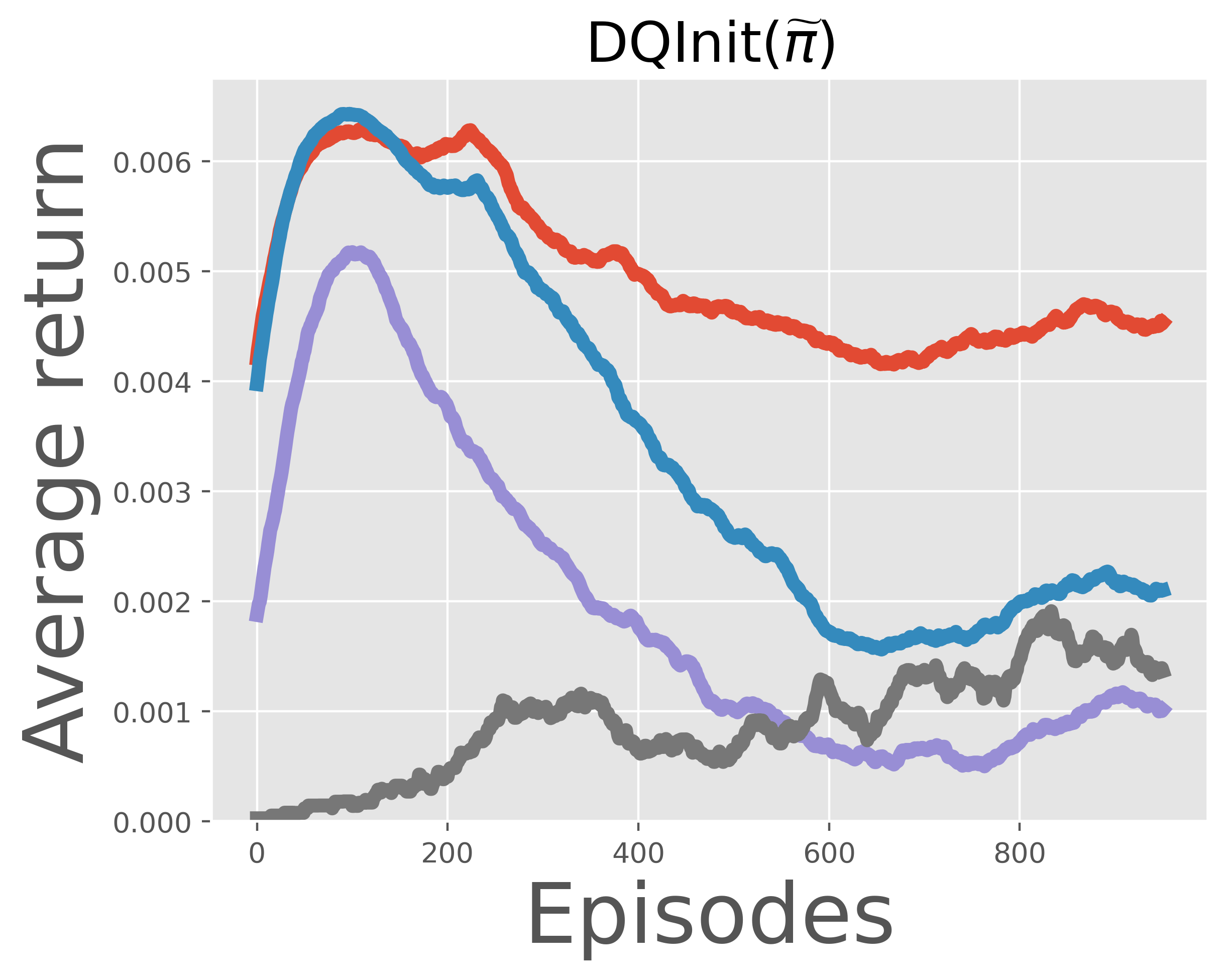}
 &\includegraphics[width=1 \linewidth]{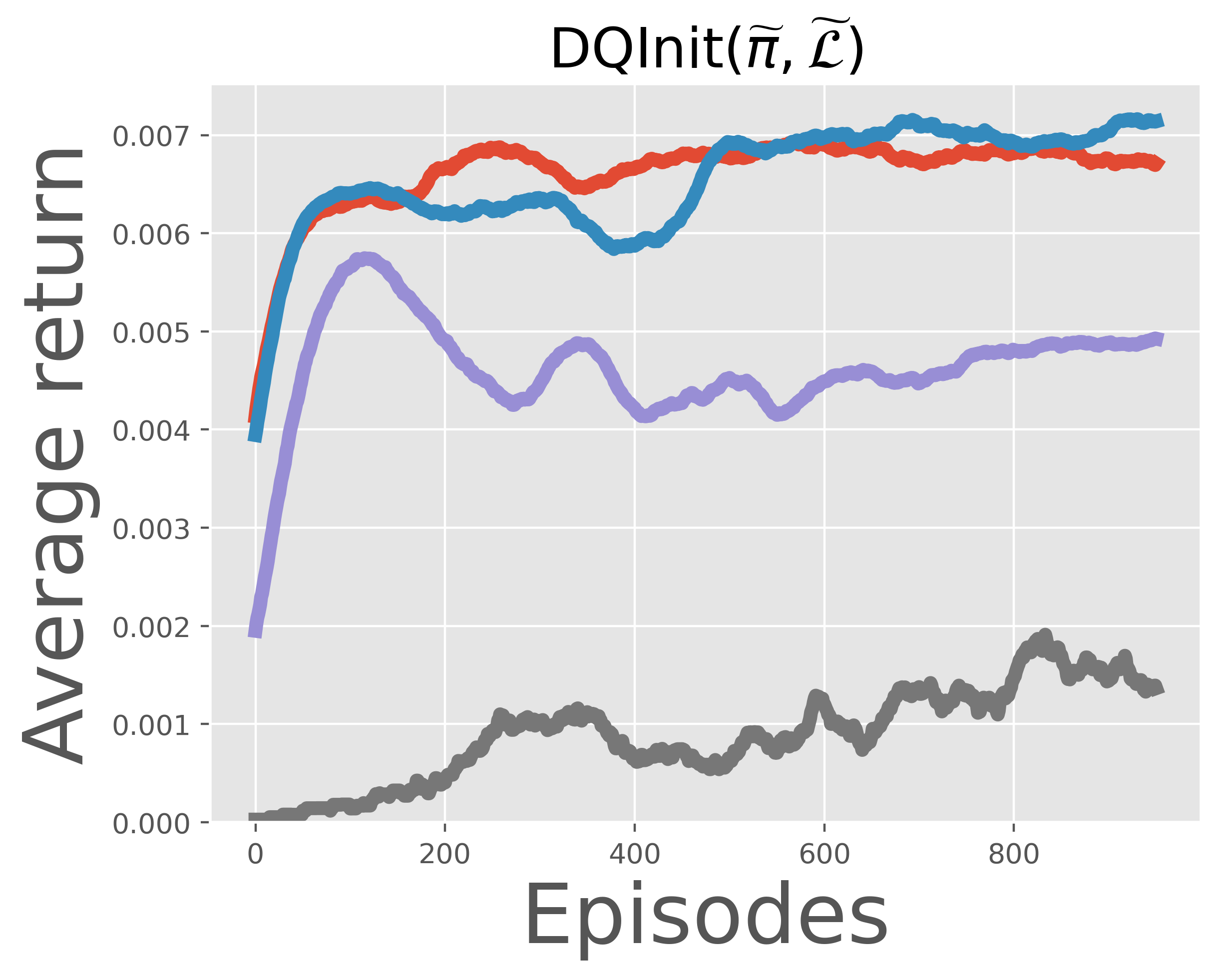}&\includegraphics[width=1 \linewidth]{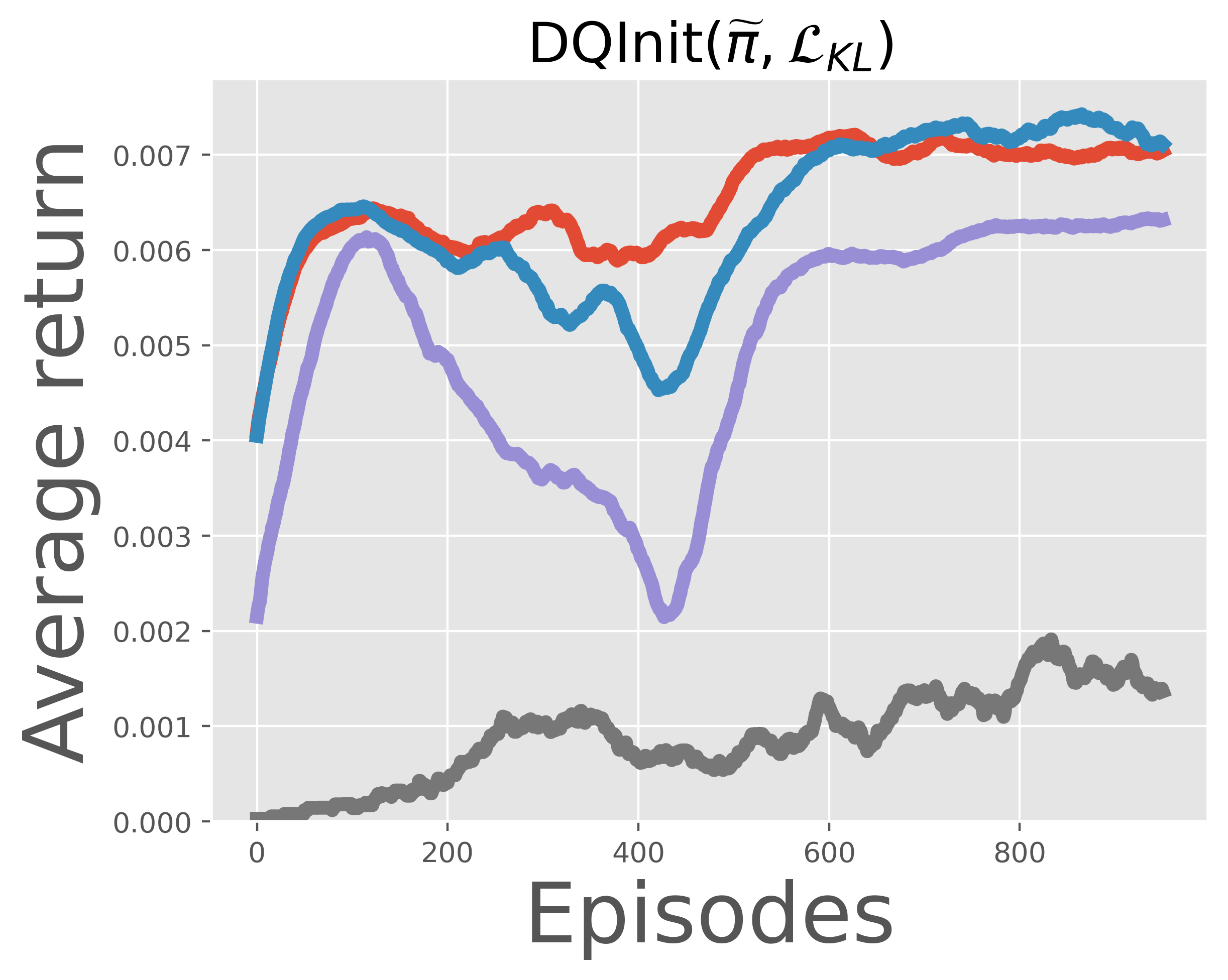}&\includegraphics[width=1 \linewidth]{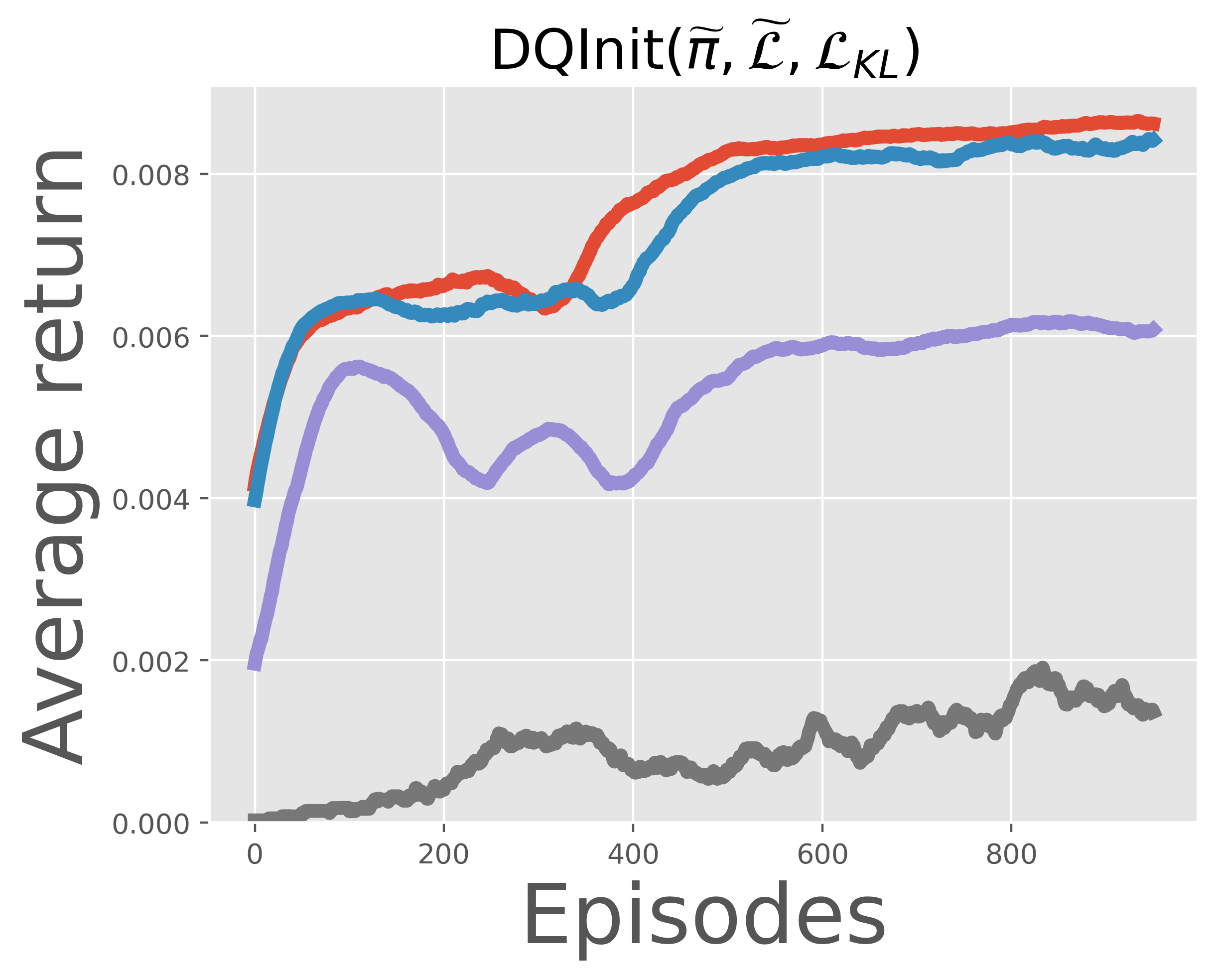} \\
\rotatebox{90}{Acrobot} &\includegraphics[width=1 \linewidth]{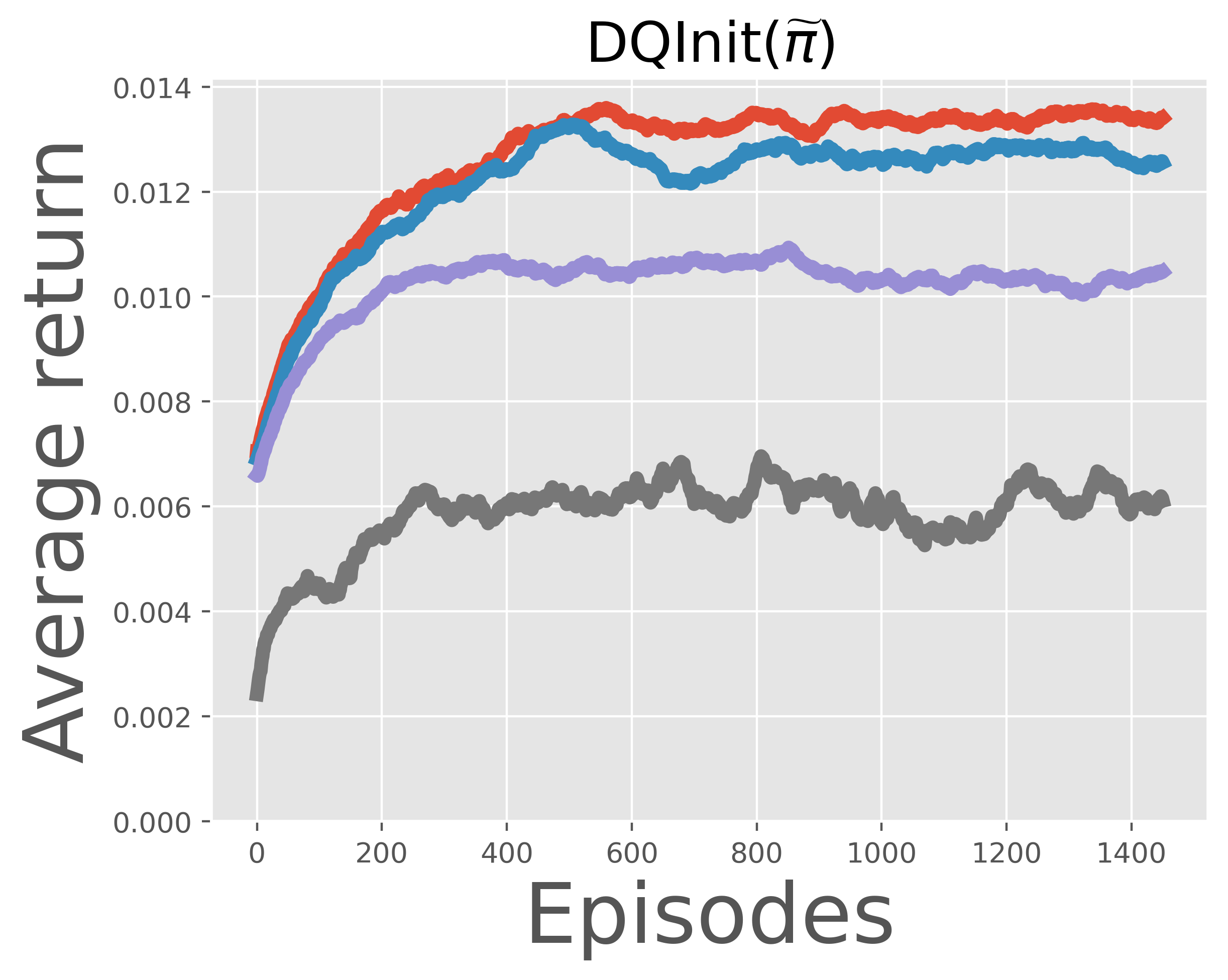}
 &\includegraphics[width=1 \linewidth]{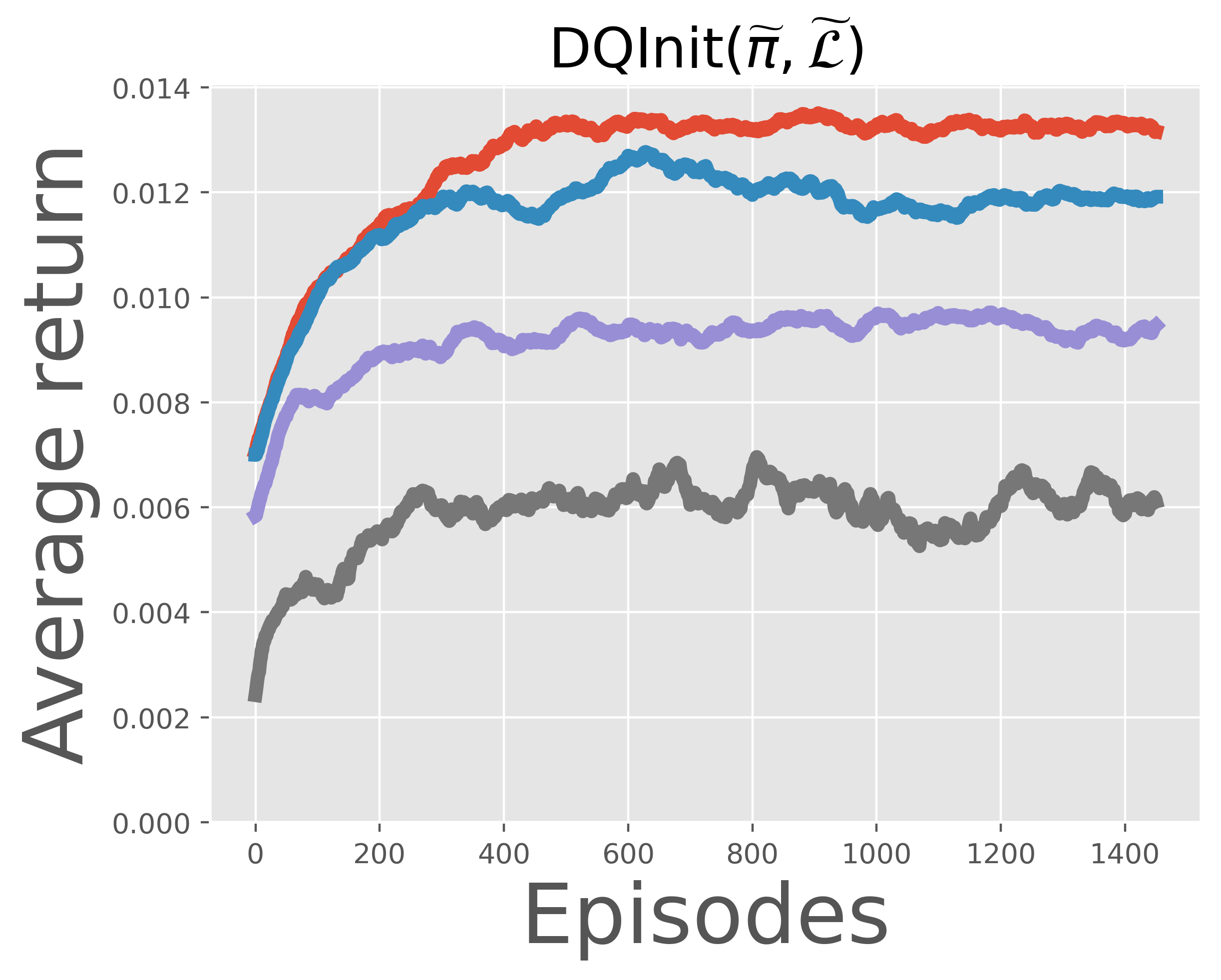}&\includegraphics[width=1 \linewidth]{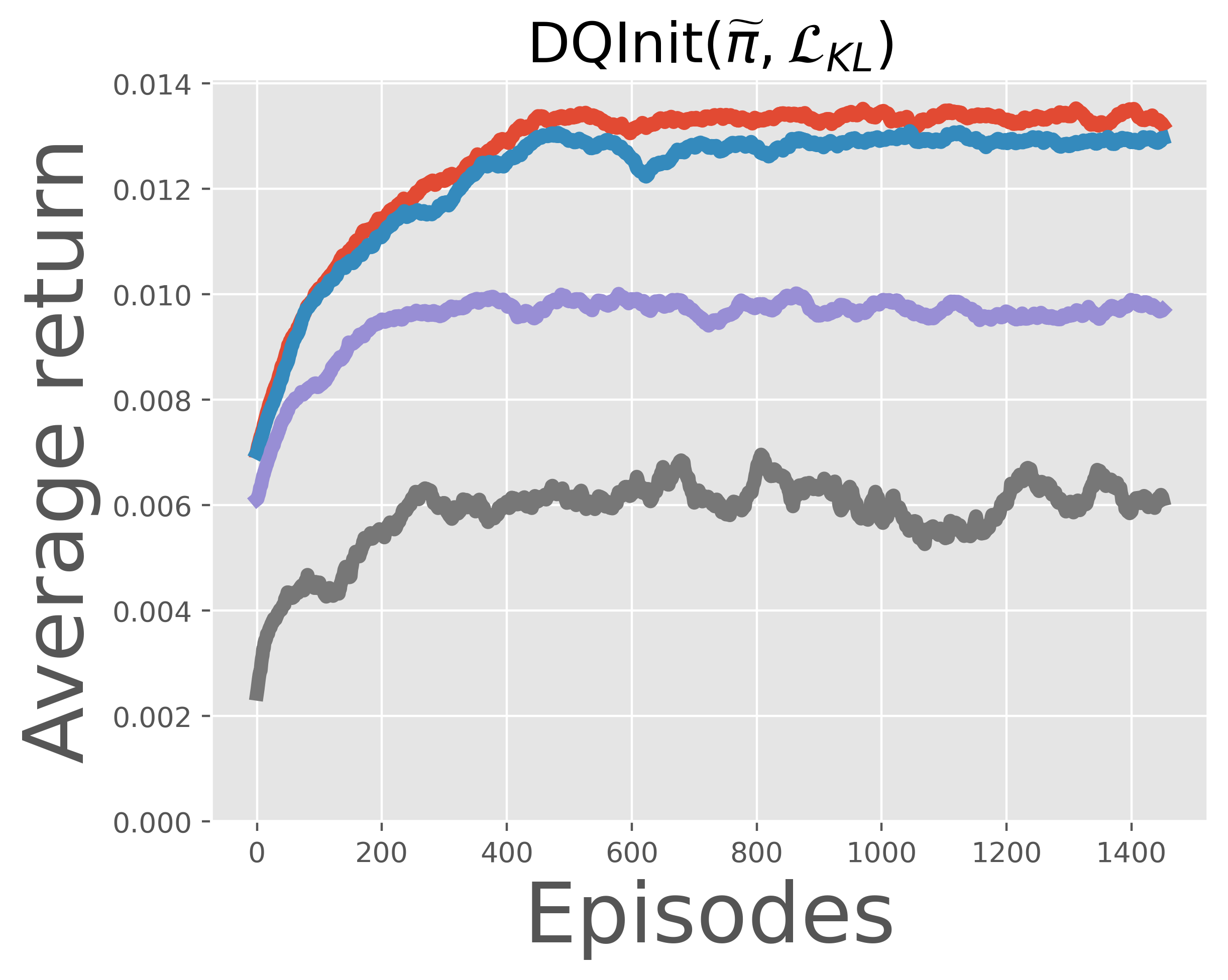}&\includegraphics[width=1 \linewidth]{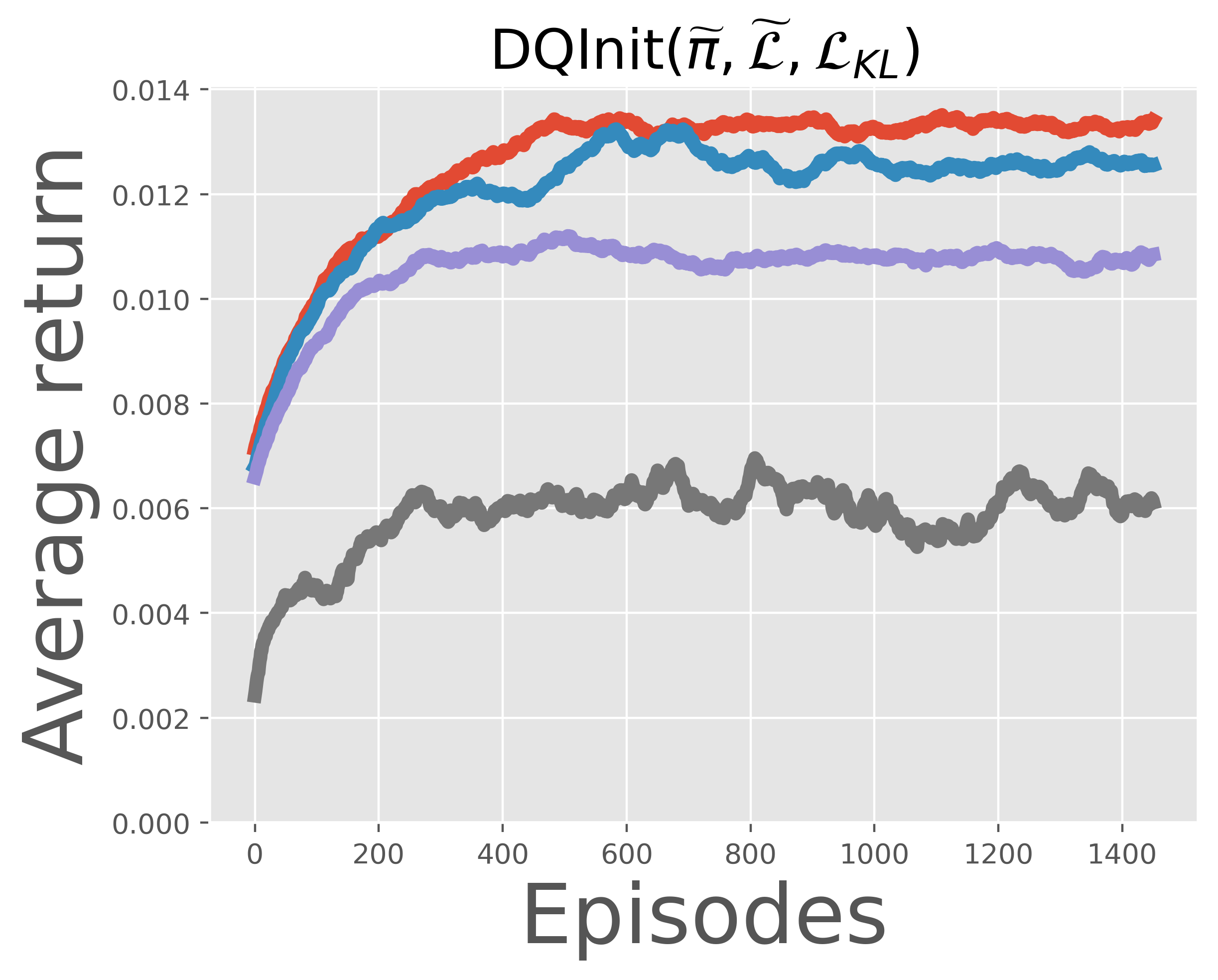} \\
\rotatebox{90}{Cartpole}&\includegraphics[width=1\linewidth]{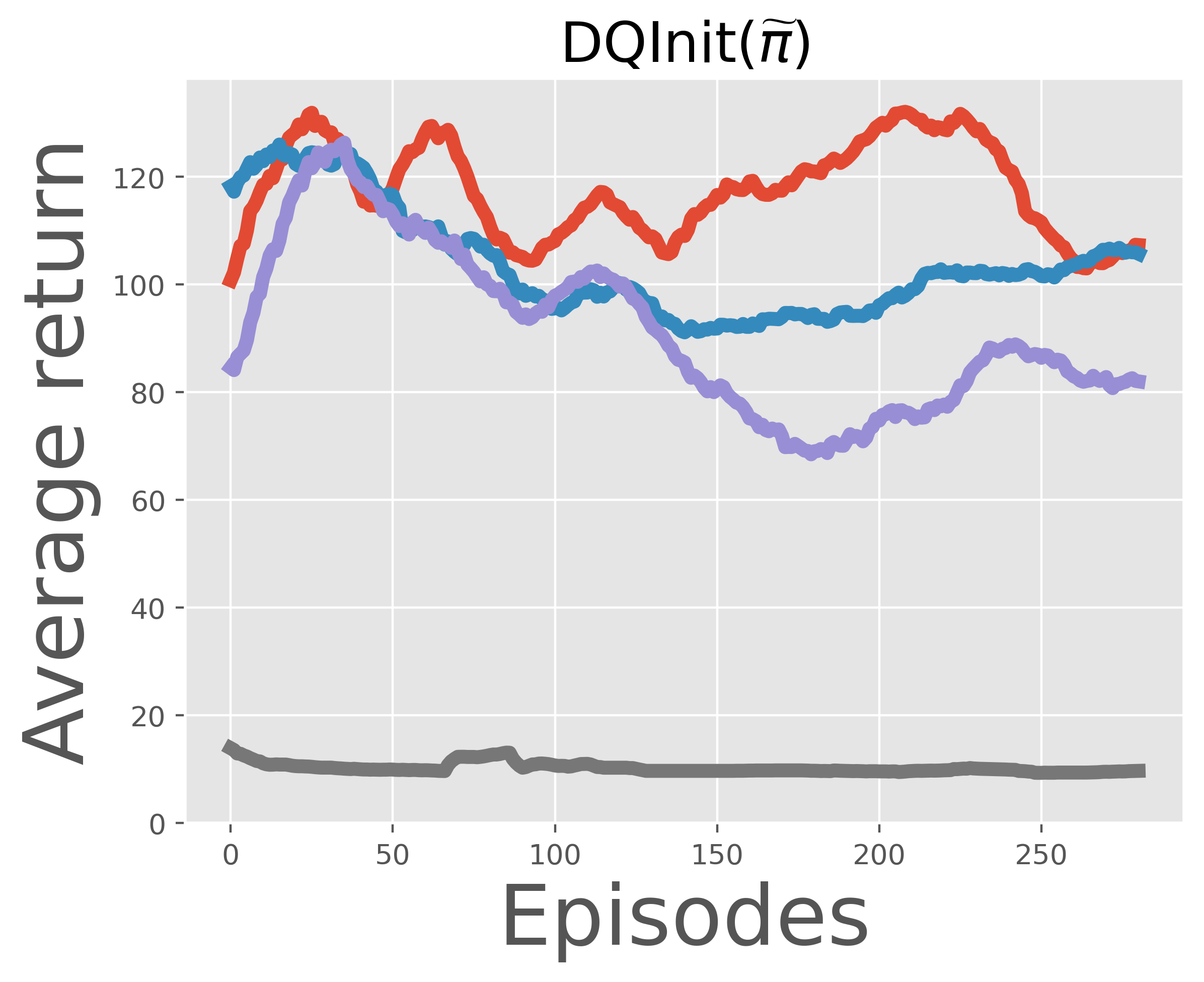}
 &\includegraphics[width=1 \linewidth]{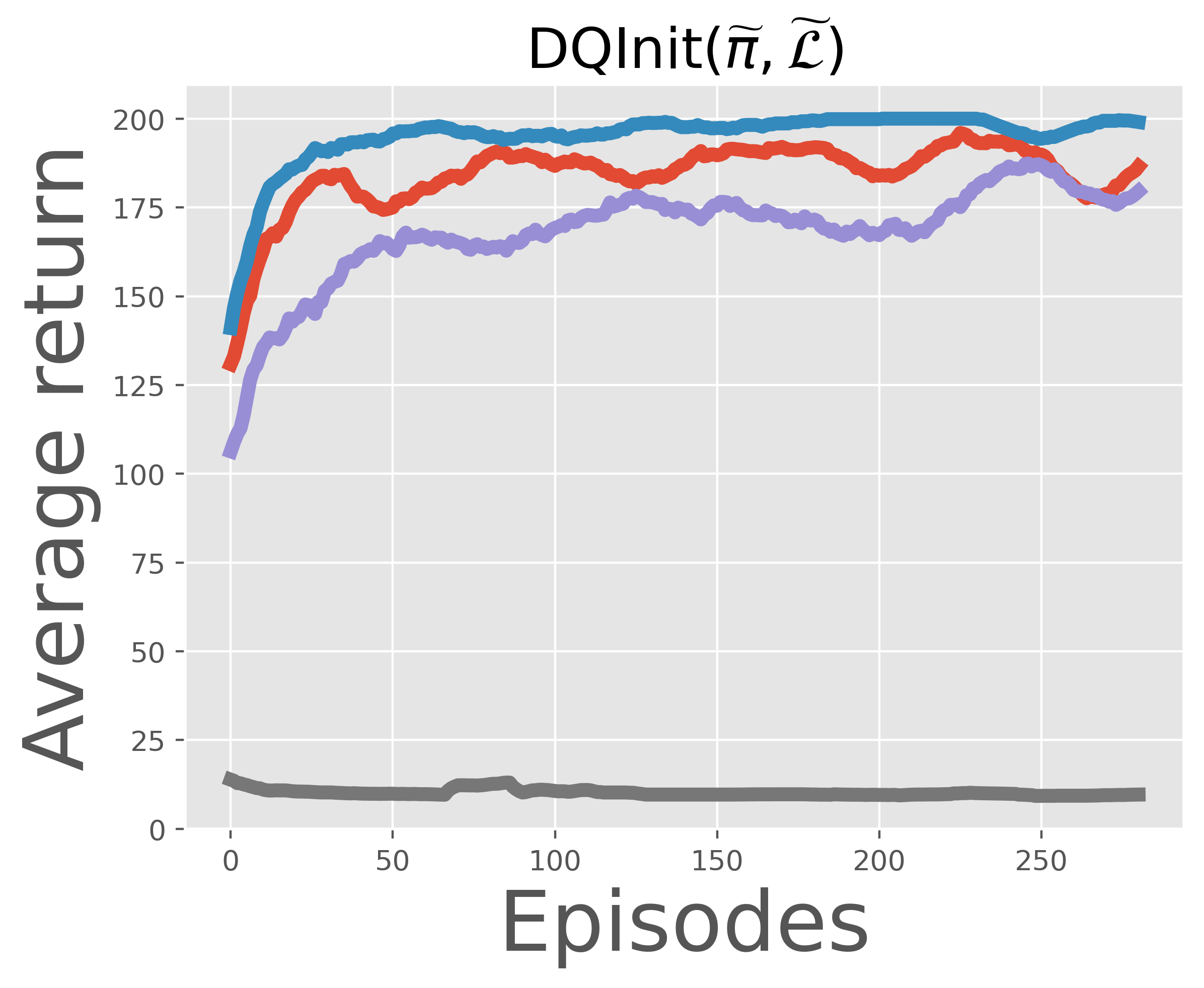}&\includegraphics[width=1 \linewidth]{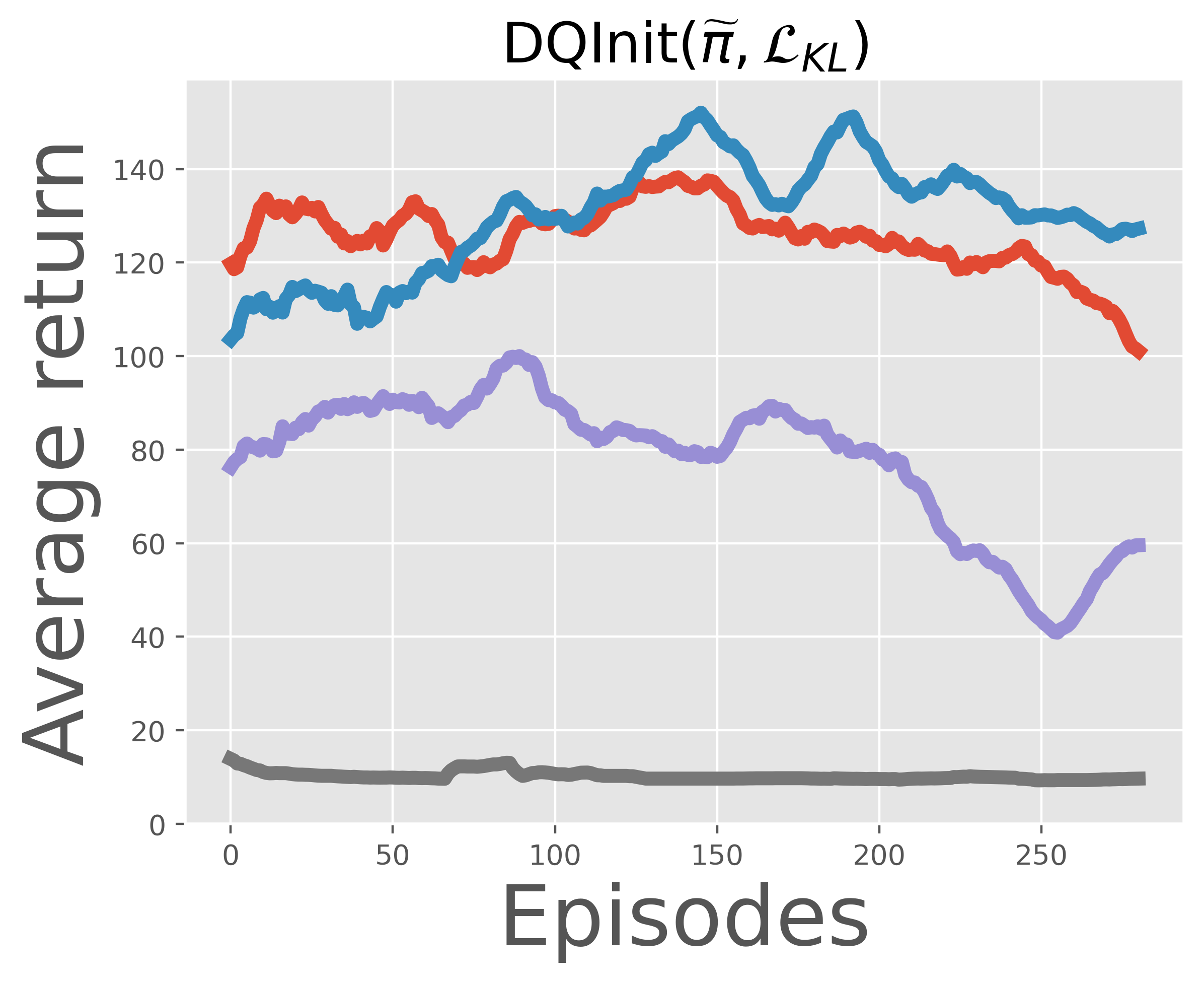}&\includegraphics[width=1 \linewidth]{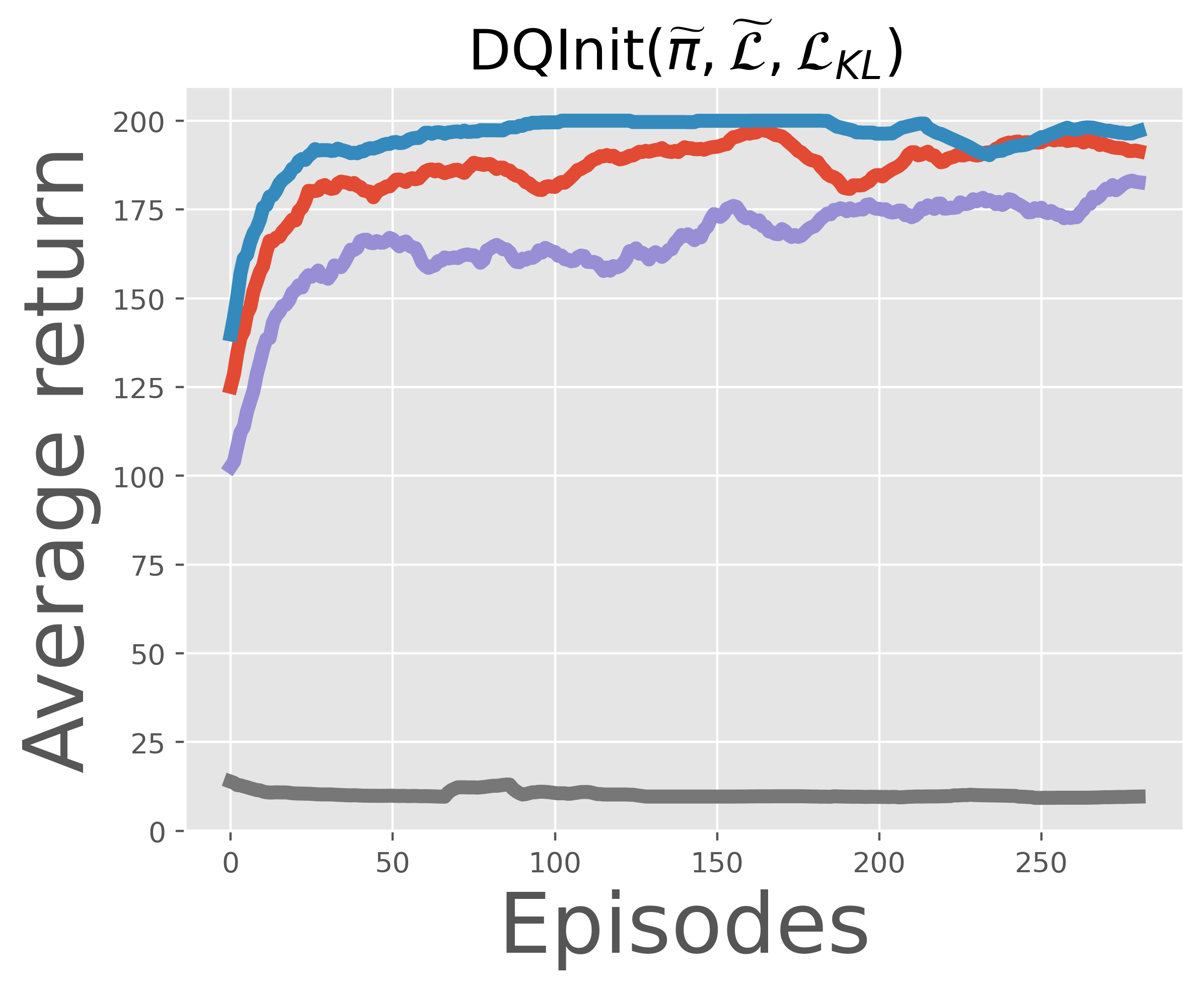} 
\end{tabular}
\includegraphics[width=.8\linewidth]{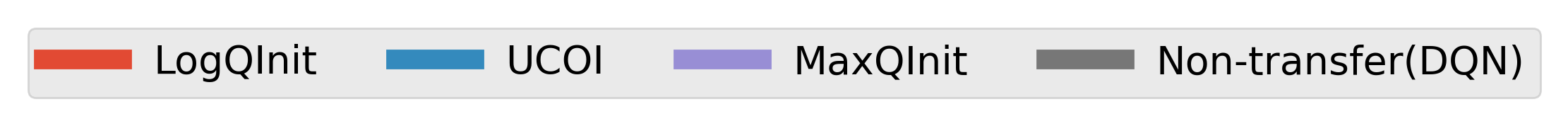}
\caption{
Performance of $\textsc{DQInit}$ across initialization strategies (MaxQInit, UCOI, LOGQInit) and usage modes ($\tilde{\pi}$, $\tilde{L}$, $L_{\text{KL}}$). LOGQInit performs best in MountainCar, UCOI in CartPole. Combining all modes yields the most stable performance across environments.}\label{fig:init}	
\end{figure*}
\subsubsection{Value initialization loss}

The flag $use_{\widetilde{\mathcal{L}}}$ defines an initialization loss to encourage the learned Q-function to match the initialized values early in training:

$$
\widetilde{\mathcal{L}} = \text{MSE}(\widetilde{Q}, Q^{\theta})
$$

This loss is inspired by the idea of initial value propagation in tabular Q-learning: As the agent encounters a state and selects an action, the corresponding Q-value is updated using TD learning. Our loss mimics this process by scaling the learned values toward the adaptive function that begins with the initialized value. As the adaptive Q-function $\widetilde{Q}$ gradually aligns with the learned Q-function $Q^{\theta}$, the loss naturally decays to zero, allowing training to rely increasingly on standard TD updates.
\begin{figure*}[h]
\centering
\subfigure[Mountain car]{\includegraphics[width=.33\linewidth]{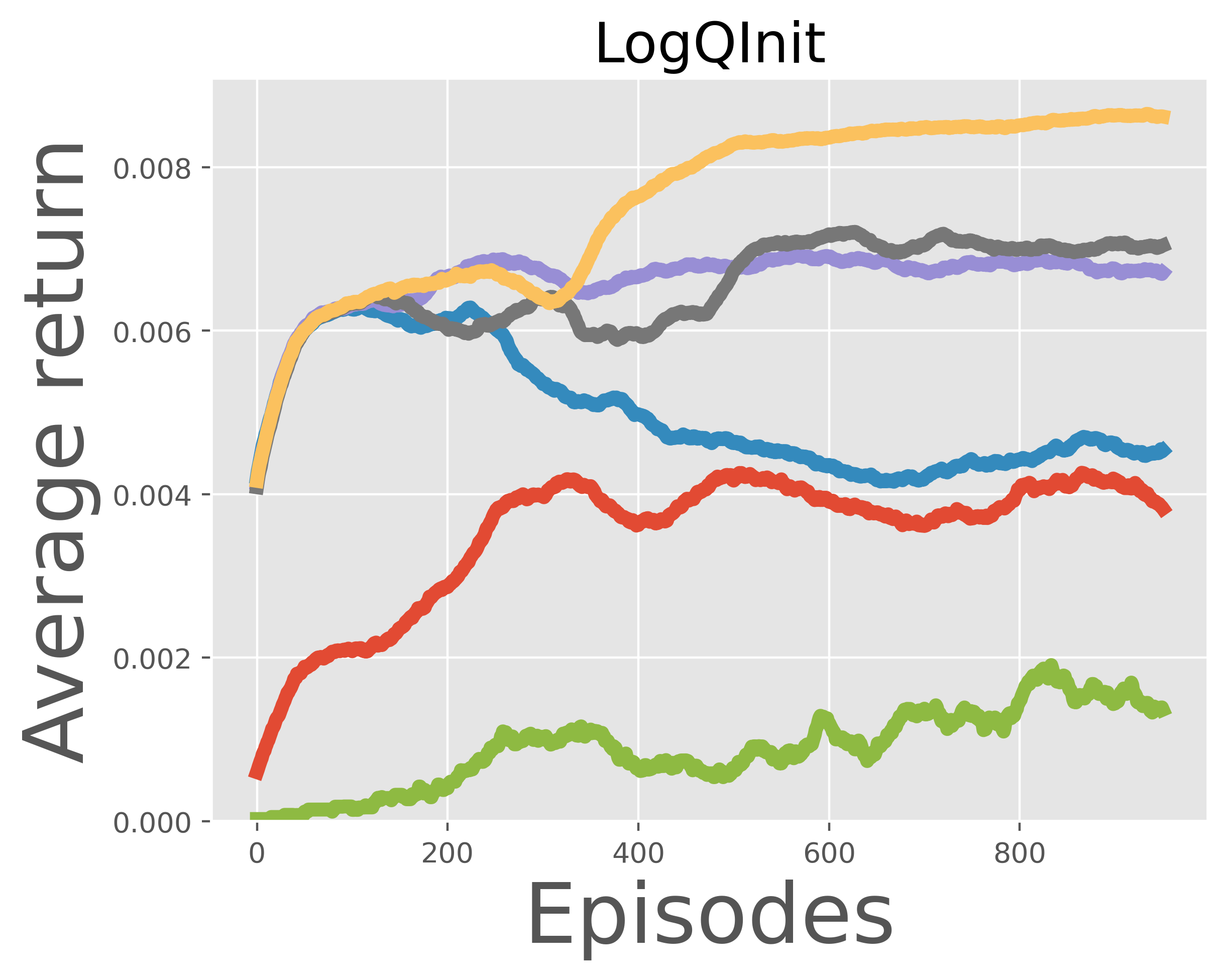}
}\subfigure[Acrobot]{\includegraphics[width=.33\linewidth]{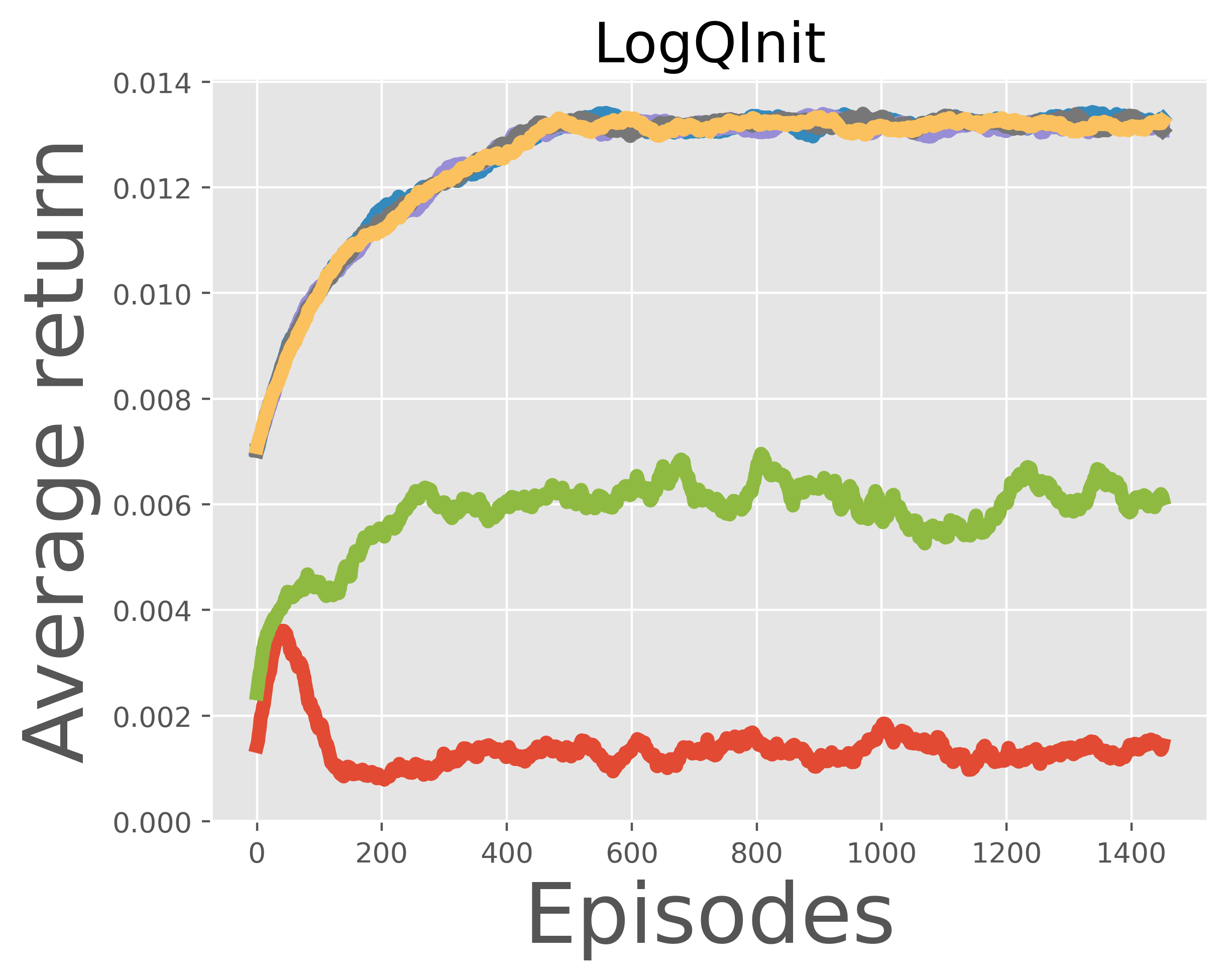}
}\subfigure[Cartpole]{\includegraphics[width=.33\linewidth]{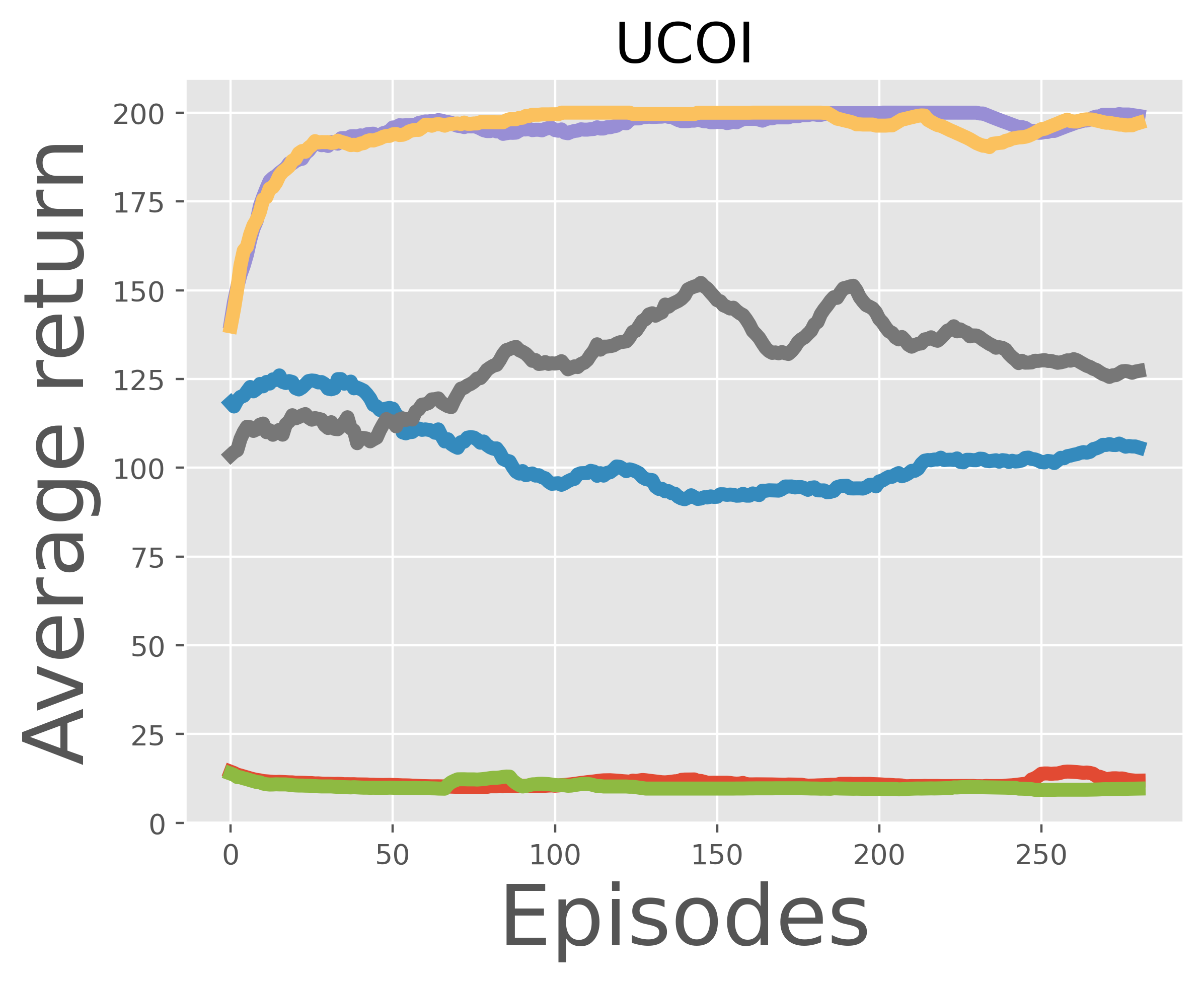}
}

\includegraphics[width=.8\linewidth]{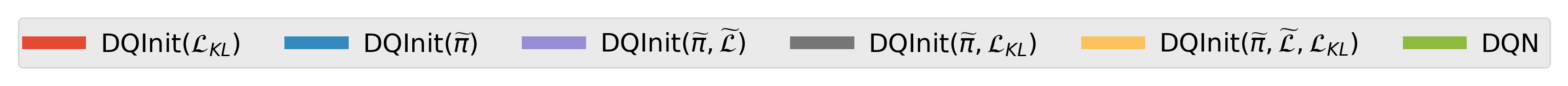}
\caption{Performance of the different $\textsc{DQInit}$ usage modes for each environment’s best-performing initialization function(LOGQInit for MountainCar and Acrobot, UCOI for CartPole). Results for other initialization strategies are provided in the supplementary materials.} \label{fig:mode}
\end{figure*}

\subsubsection{Policy Distillation Loss}

We also explore a distillation-based mode, where the agent aligns its learned policy with source models. However, since $\widetilde{Q}$ evolves (due to knownness), it is not suitable as a distillation target. Instead, we compute a KL divergence between the agent’s policy derived from $Q^{\theta}$ and the fixed initialization function $Q^{\emptyset}$:
$$
\mathcal{L}_{KL} = \text{KL}(\pi^{\theta} \| \pi^{\emptyset})
$$

This usage mirrors traditional policy distillation, with the key difference that the source policy is derived from a Q-table instead of a teacher model.

\section{Experiments}

\subsection{Task Setting}
We evaluate on three classical control environments, each varied across a distribution of tasks by modifying physical parameters that affect dynamics as follows:
\begin{itemize}
\item MountainCar: A sparse-reward environment with deterministic transitions. We variation to the dynamics by adding a Gaussian noise $\eta$ into the velocity updates such as $\eta \sim \mathcal{N}(0, 0.02)$.

\item Acrobot: A less sparse domain with stochastic next states due to its nonlinear dynamics. Task variation is created by sampling the lengths of the two links from $\mathcal{N}(0.95, 0.1)$, clipped to $[0.7, 1.2]$.

\item CartPole: A dense-reward, horizon-based task. We vary the pole length using $\mathcal{N}(0.5, 0.2)$, clipped to $[0.2, 1.2]$, which affects return magnitudes and early termination likelihood.
\end{itemize}

\subsection{Knowledge Base Preparation}

We construct the knowledge base by learning 30 tasks per environment and saving their DQN models and the corresponding Q-table. To train the tabular value functions used for initialization, we adopt a binary reward scheme in MountainCar and Acrobot: a reward of 1 is given upon success, and 0 otherwise. This improves convergence and amplifies value differences in the Q-table, compared to the original sparse negative rewards. During transfer, the binary reward is also used in the target tasks when learning with DQN. For CartPole, both the Q-table and DQN are trained using the original dense rewards, which are already informative. Full details about the hyperparameters are listed in Table 1 in the supplementary materials.

\subsection{Initialization strategies:}
We test three VFI methods proposed in tabular RL:
\begin{itemize}
    \item \textsc{MaxQInit}~\cite{maxqinit}: Initializes using the maximum Q-value across prior tasks, ensuring optimism under uniform task distributions.
    \item $\mathsf{UCOI}$~\cite{mehimeh2023value}: Balances optimistic initialization with the empirical mean, modulated by PAC-based confidence and task uncertainty.
    \item \textsc{LogQInit}~\cite{mehimeh2025}: Assumes log-normality in value distributions, using the log-mean to initialize Q-values.
\end{itemize}

\subsection{$\textsc{DQInit}$ modes:} For each strategy, we tested different "modes" of $\textsc{DQInit}$ by activating only one flag, or two, or three together, and we get the following modes:  
($\widetilde{\pi}$), ($\widetilde{\pi},\widetilde {\mathcal{L}}$), 
($\widetilde{\pi},\mathcal{L}_{KL}$), ($\widetilde{\pi},\widetilde{\mathcal{L}},\mathcal{L}_{KL}$).
\subsection{Baseline}
We compare $\textsc{DQInit}$ with the standard policy distillation method, where the target policy is derived by averaging the outputs of multiple expert models. Additionally, we include a comparison with Jump-Start Reinforcement Learning (JSRL), in which a decay factor determines the probability of selecting actions from the expert policy versus the current task policy.

\section{Results and discussion}

Our experimental objective is to investigate \textsc{DQInit} by addressing key analytical questions, alongside validating its effectiveness against basic transfer baselines. The results and discussion are structured around the following guiding questions:

\begin{enumerate}
\item To validate whether VFI can generalize to DRL settings with continuous state spaces.
\item To assess the flexibility of our approach across three usage modes: soft policy guidance, value initialization loss, and policy distillation.
\item To test our hypothesis that tabular value functions offer greater robustness and scalability as a source of knowledge compared to model-based Q-networks.
\item To position the performance of \textsc{DQInit} across the baselines.
\end{enumerate}
For all experiments, the DQN+$\textsc{DQInit}$ was trained on 30 tasks from a similar distribution as knowledge source tasks. All figures report the average return across these 30 tasks.
\subsection{Initialization Strategies Comparison}
\noindent \textbf{Summary of Literature Results in Tabular VFI:} As stated in the main paper, one of our primary objectives is to evaluate whether the conclusions drawn from value function initialization (VFI) in tabular reinforcement learning (RL) hold in the deep RL setting. Below, we summarize key observations from prior tabular VFI studies:

\begin{itemize}
    \item \textbf{Observation 1:} In the MaxQInit method, initializing with the maximum value across previous tasks consistently improves jumpstart performance when tasks are sampled from a uniform distribution.

    \item \textbf{Observation 2:} In the UCOI study, experiments on both uniform and skewed task distributions showed that UCOI outperforms MaxQInit in non-uniform settings. As task variation becomes closer to uniform, the performance gap between UCOI and MaxQInit narrows.

    \item \textbf{Observation 3:} In the LOGQInit work, it was found that if task difficulty (e.g., path length to goal in sparse-reward settings) follows a normal distribution, the resulting value function follows a log-normal distribution. Consequently, in such settings, the LOGQInit method outperforms UCOI and MaxQInit. The larger the variance in task difficulty, the greater the performance gap. However, these results were primarily derived from sparse-reward environments and may not directly apply to dense-reward tasks.

    \item \textbf{Observation 4:} Across studies, it was generally observed that the smaller the task variance, the closer the performance between different initialization methods.
\end{itemize}

We now relate these findings to our results in DRL: Figure~\ref{fig:init} presents the performance comparison of initialization strategies. The results are consistent with theoretical expectations: all methods improve jumpstart performance. In sparse-reward environments like MountainCar, \textsc{LogQInit} performs robustly, supporting the theoretical findings discussed in previous chapters. For CartPole, $\mathsf{UCOI}$ outperforms others—likely due to its ability to identify confident regions in finite-horizon tasks with penalty-based terminations, where \textsc{LogQInit}'s log-normality assumption may not hold. In Acrobot, all methods show similar performance, potentially due to low task variance under the applied dynamics perturbation. These results confirm that VFI in DQN is both feasible and influential, with different strategies leading to distinct learning behaviors.

\begin{itemize}
    \item \textbf{MountainCar:} As a sparse-reward environment with path-length-based difficulty, this aligns with Observation 3. Our results showed a consistent hierarchy: LOGQInit $>$ UCOI $>$ MaxQInit.

    \item \textbf{Acrobot:} the variation in dynamics for this environment was small, resembling a near-uniform distribution. As per Observation 4, the performance differences between methods were minimal. Nonetheless, LOGQInit slightly outperformed others, consistent with mild log-normal value behavior.

    \item \textbf{CartPole:} This dense-reward, horizon-based task diverges from the assumptions in Observation 3. However, Observations 1 and 2 remain relevant: UCOI outperformed MaxQInit, reflecting non-uniformity in dynamics. LOGQInit did not provide a clear advantage, consistent with the absence of sparse reward structure.
\end{itemize}

\textbf{Conclusion:} Our experimental findings in DRL mirror the theoretical and empirical results observed in tabular VFI. This confirms that our proposed method is not a naive extension but a grounded adaptation. It also opens the door to further theoretical investigations on how value distributions behave across task distribution, and how this can guide principled strategies for probabilistic initialization in DRL.

\subsection{Modes performance}
Next, to further analyze the contribution of each usage mode, we fix the best-performing initialization strategy per environment (LOGQInit for MountainCar and Acrobot, UCOI for CartPole) and compare different combinations of $\textsc{DQInit}$ modes in Figure \ref{fig:mode}. Soft policy guidance alone ($\textsc{DQInit}(\widetilde\pi)$) provides strong early-stage performance but often decreases as training progresses. In contrast, adding the value loss ($\textsc{DQInit}(\widetilde\pi,\widetilde{\mathcal{L}})$) leads to better long-term results by encouraging early alignment between the learned Q-function and the initialized values. Combining soft advice with policy distillation ($\textsc{DQInit}(\widetilde\pi,\mathcal{L}_{K})$) produces mixed results: it works well in MountainCar and Acrobot but performs poorly in CartPole. This may be due to CartPole's higher value magnitudes, which can amplify KL divergence effects and destabilize learning. Notably, combining all three usage modes ($\textsc{DQInit}(\widetilde\pi,\widetilde{\mathcal{L}},\mathcal{L}_{KL})$) yields the most robust and stable performance across all environments: the agent effectively boosts its jumpstart, approximates the initial value, and learn the teacher's behavior.

\begin{figure}[htp!]
\subfigure[Mountain car]{
\includegraphics[width=.3\linewidth]{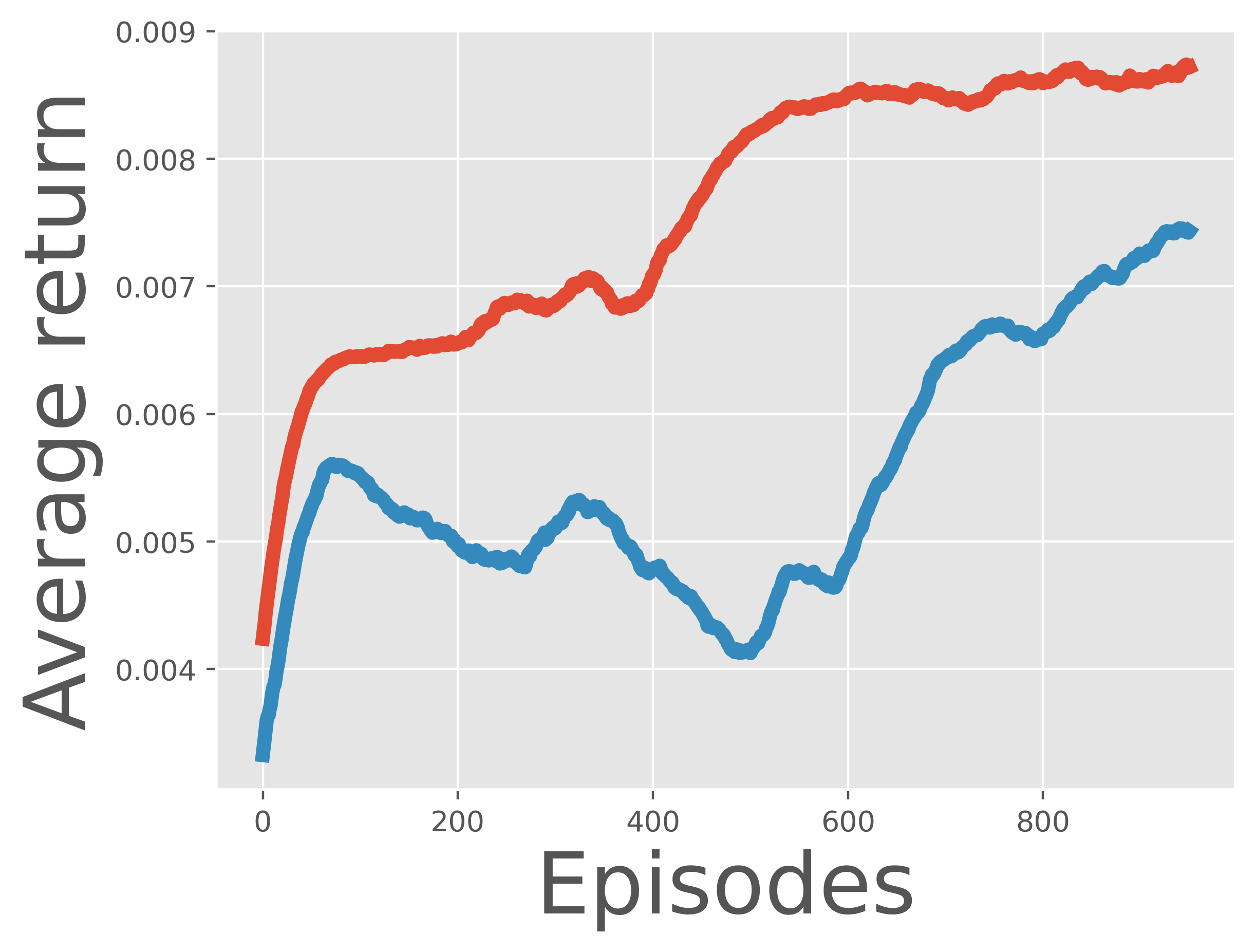}
}
\subfigure[Acrobot]{
\includegraphics[width=.3\linewidth]{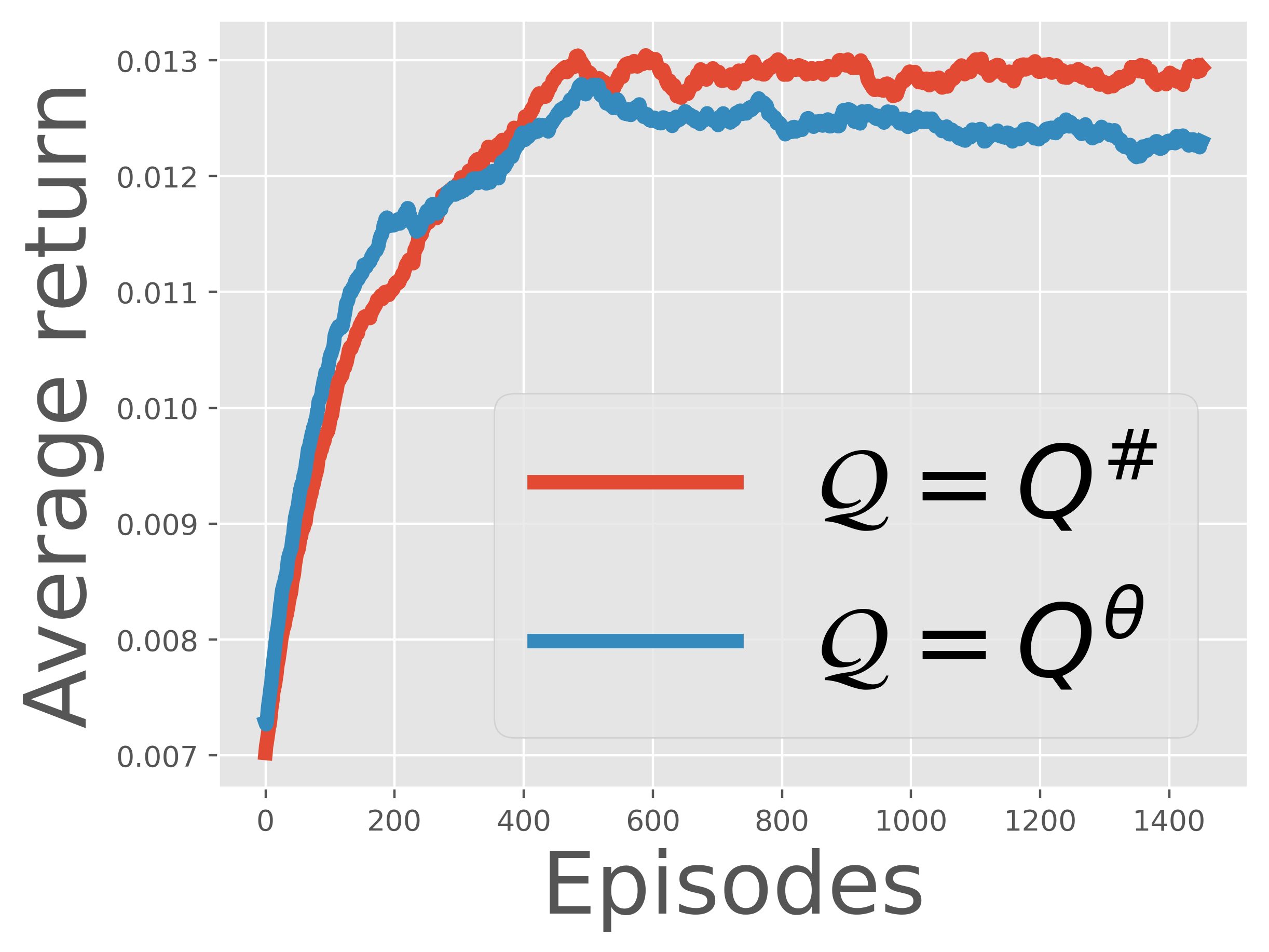}
}
\subfigure[Cartpole]{
\includegraphics[width=.3\linewidth]{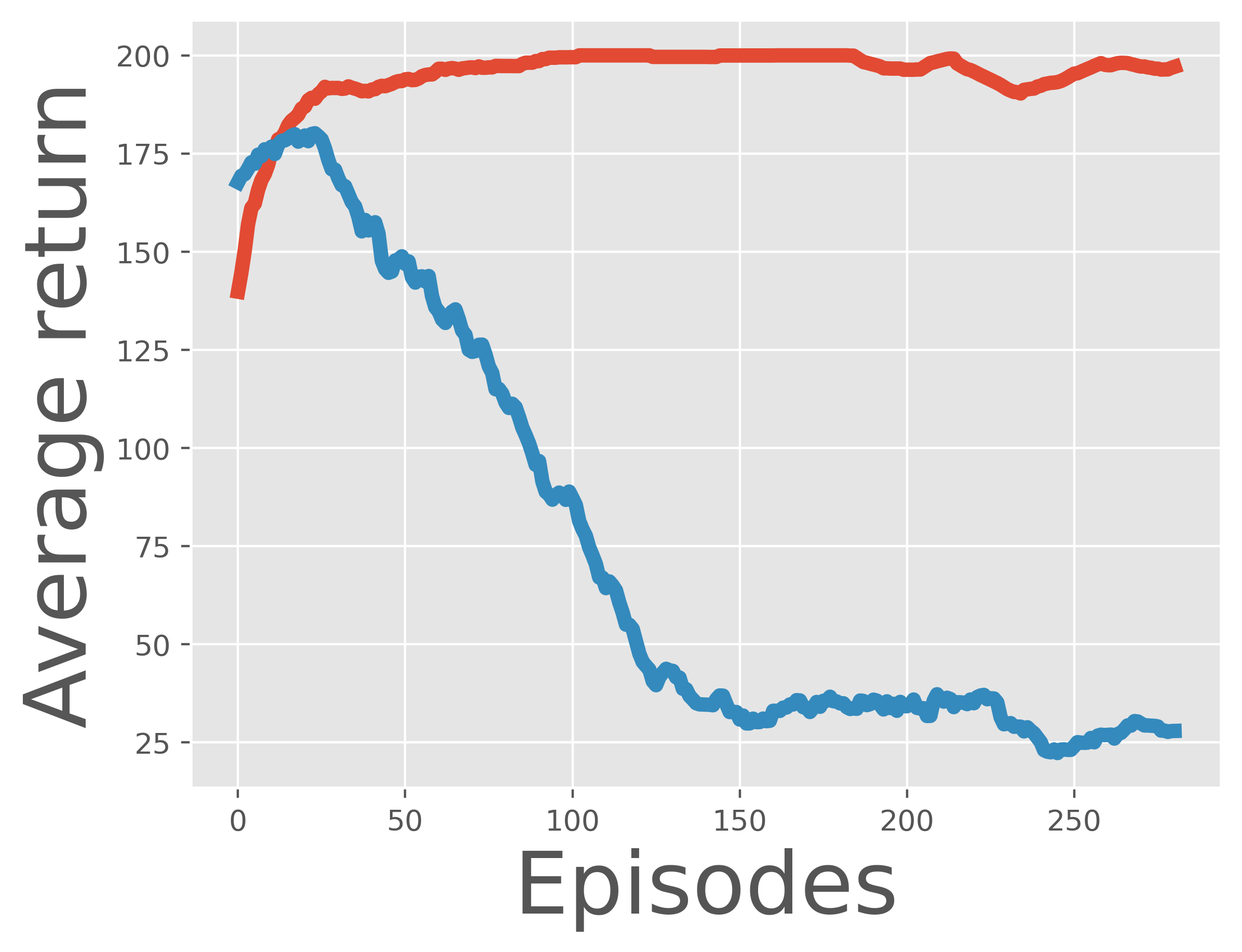}
}
\caption{Performance of $\textsc{DQInit}(\widetilde\pi,\widetilde{\mathcal{L}},\mathcal{L}_{KL})$  initialized with different source networks output ($\mathcal{Q}=Q^{\theta}$) vs tabular value function as knowledge base ($\mathcal{Q}=Q^{\#}$)}\label{fig:tab}
\end{figure}

\subsection{Source value function:} To evaluate the effectiveness of using tabular initialization in \textsc{DQInit} compared to directly using the raw output of trained models, we conduct a comparative experiment using the best performing mode of each environment $\textsc{DQInit}(\widetilde\pi,\widetilde{\mathcal{L}},\mathcal{L}_{KL})$. The results, presented in Figure~\ref{fig:tab}, demonstrate the robustness of using tabular value sources over neural network outputs for initialization. Learning from tabular source tasks yields performance that is comparable to—or even better than—initialization based on network models. This not only reduces sensitivity to the variability of neural value functions but also improves storage efficiency, as there is no need to retain full models to achieve similar or better performance.
\begin{figure*}
\centering
\subfigure[Mountain car]{
\includegraphics[width=.3\linewidth]{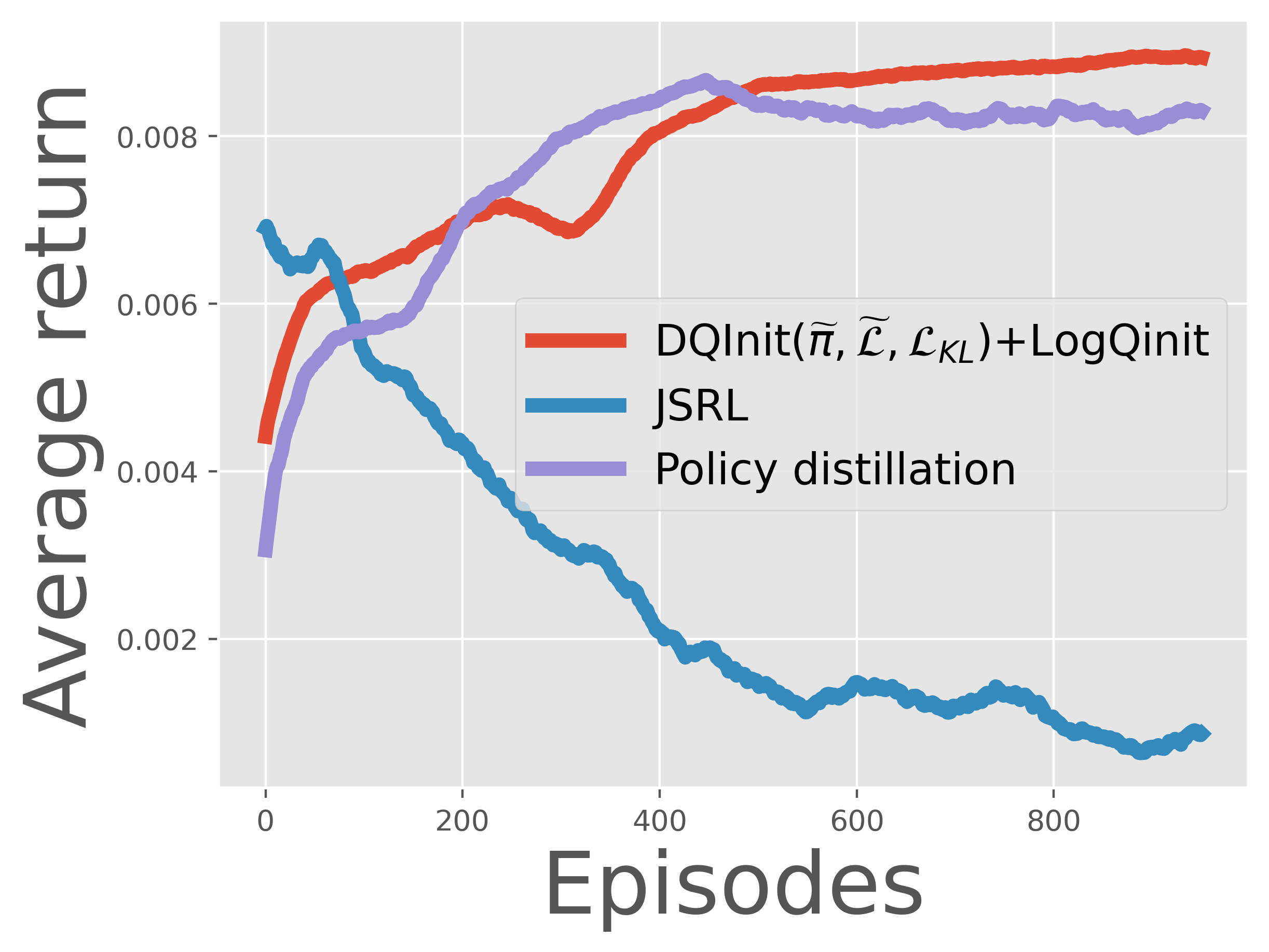}
}\subfigure[Acrobot]{
\includegraphics[width=.3\linewidth]{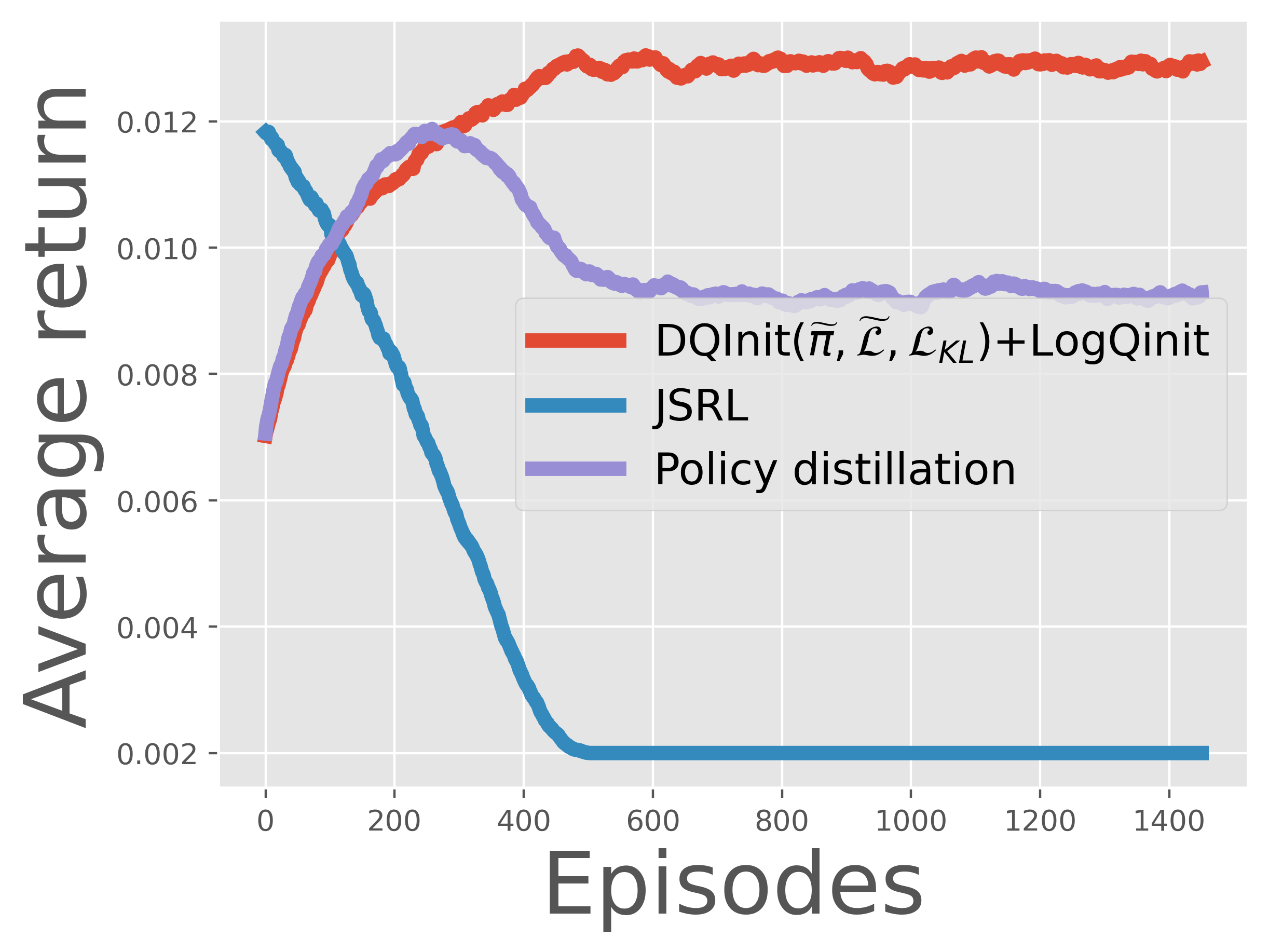}
}\subfigure[Cartpole]{
\includegraphics[width=.3\linewidth]{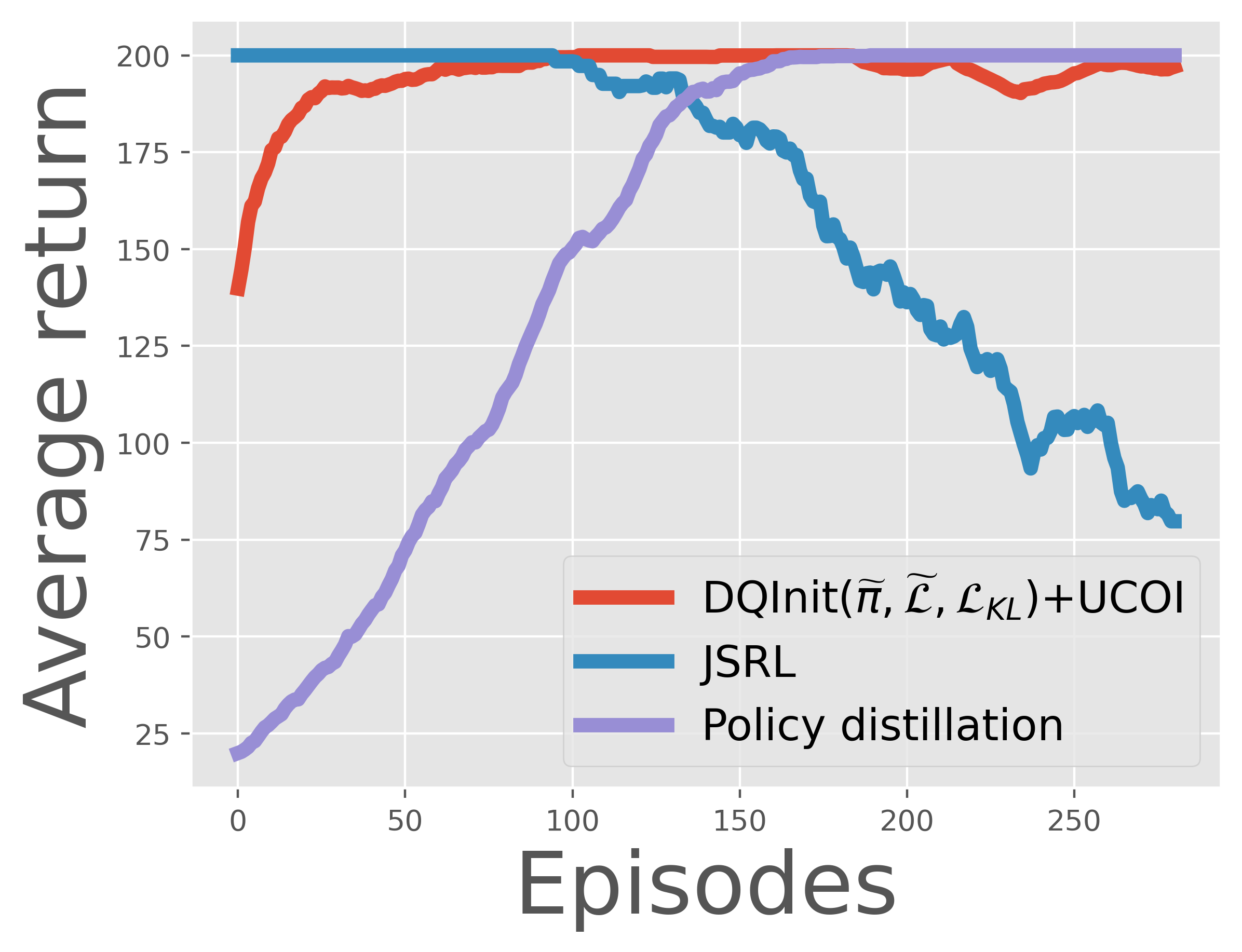}
}
\caption{Average overall reward of $\textsc{DQInit}(\widetilde\pi,\widetilde{\mathcal{L}},\mathcal{L}_{KL})$ vs  $\theta$-dependent average return across different environments }\label{fig:base}
\end{figure*}
\subsection{\textsc{DQInit} vs baseline}

Figure~\ref{fig:base} compares our full $\textsc{DQInit}$ configuration ($\widetilde\pi, \widetilde{\mathcal{L}}, \mathcal{L}_{\text{KL}}$) against two baselines: JSRL and policy distillation.  JSRL benefits from early expert guidance and achieves better initial performance than $\textsc{DQInit}$ in CartPole; however, its effectiveness diminishes over time due to the decaying influence of the expert policy.

Policy distillation exhibits varying behaviors: in MountainCar, it starts well and converges similarly to $\textsc{DQInit}$; in Acrobot, it starts well but converges to a suboptimal policy; and in CartPole, it suffers from poor initial performance and slow convergence. In contrast, $\textsc{DQInit}$ demonstrates robust performance across all tasks, combining the strengths of both jumpstart and long-term adaptability. These results position $\textsc{DQInit}$ as a promising and effective method for value-based transfer in reinforcement learning.

\begin{figure}[h]
\subfigure[Mountain car]{
\includegraphics[width=.3\linewidth]{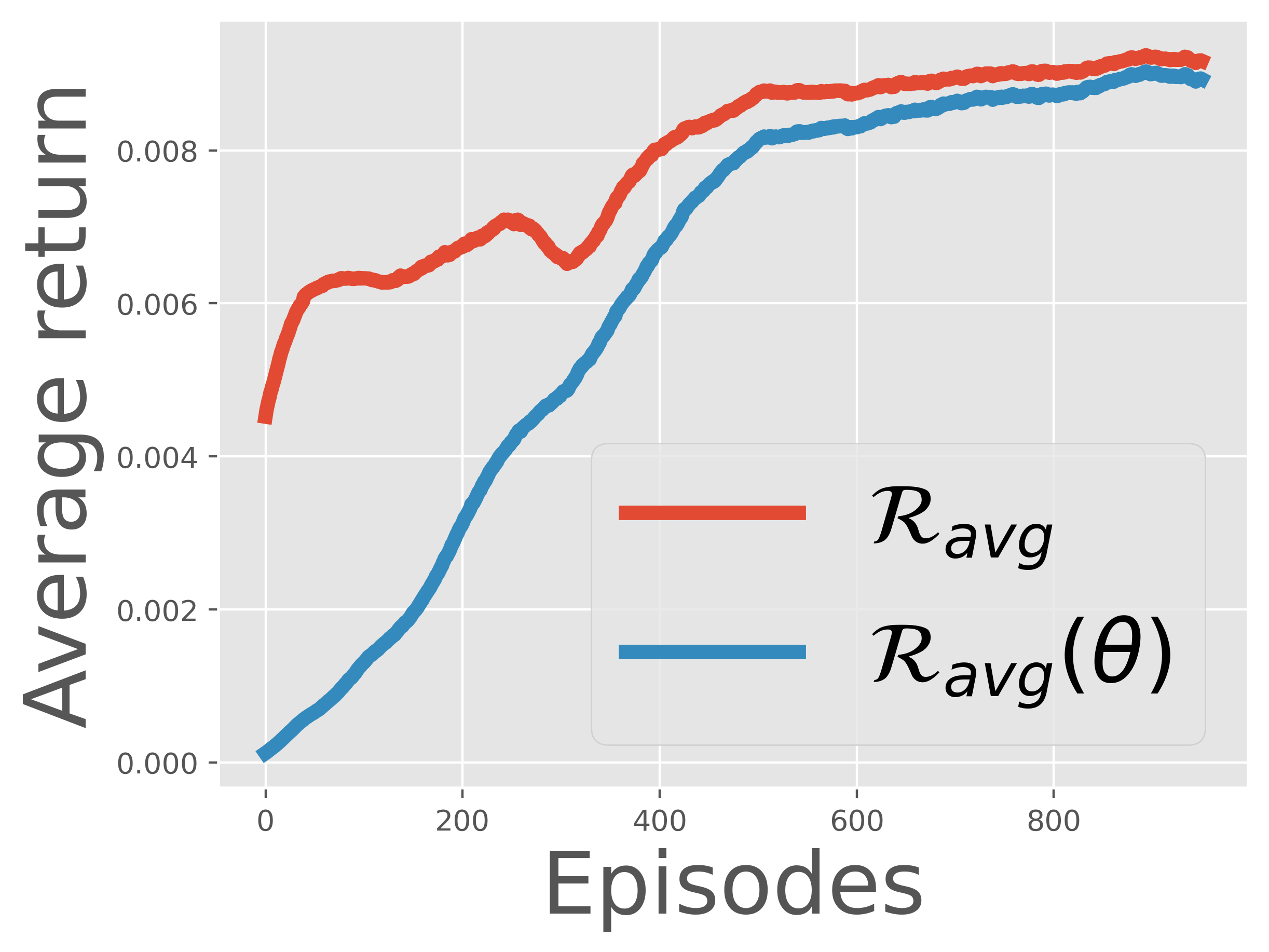}
}
\subfigure[Acrobot]{
\includegraphics[width=.3\linewidth]{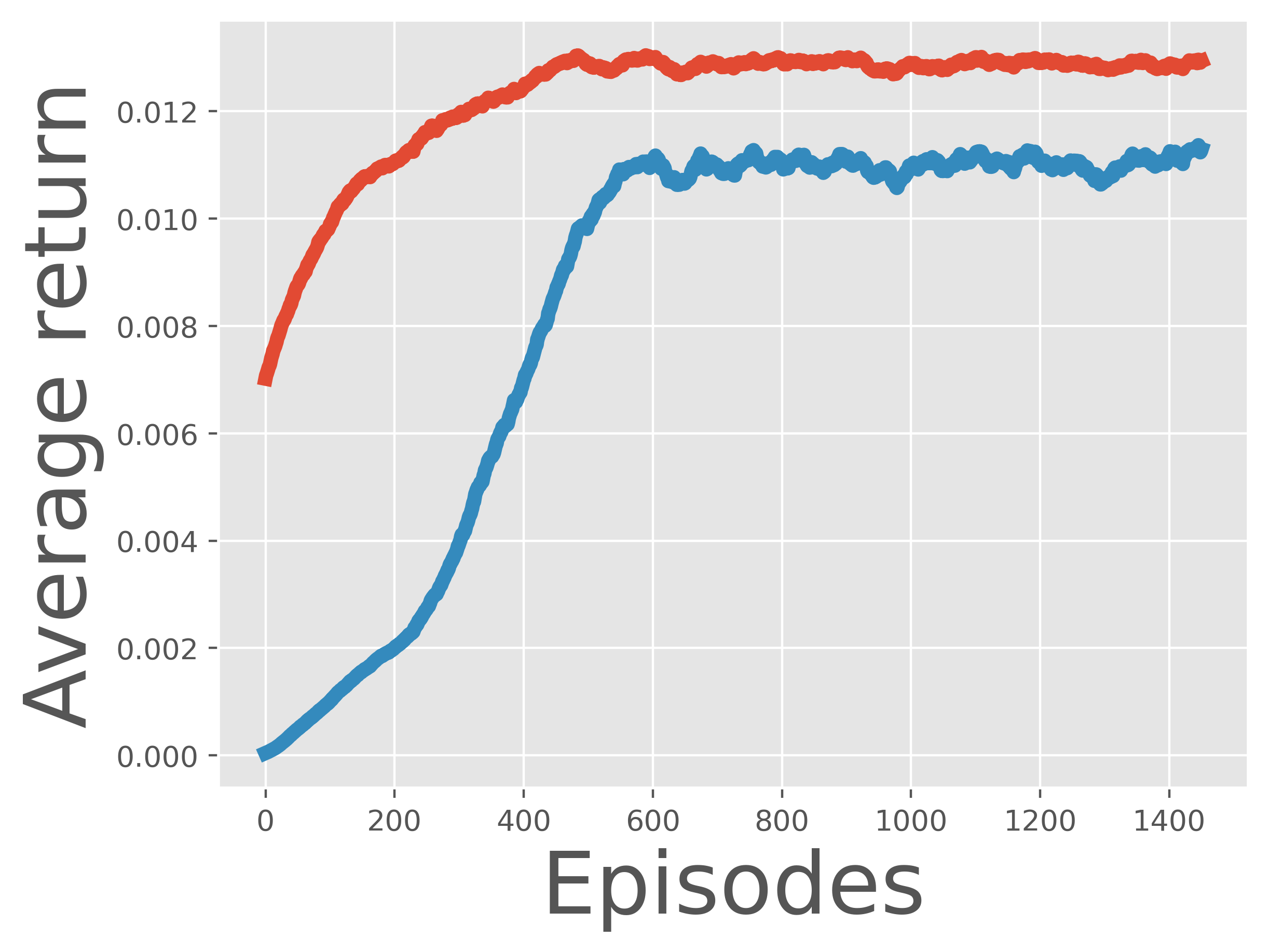}
}
\subfigure[Cartpole]{
\includegraphics[width=.3\linewidth]{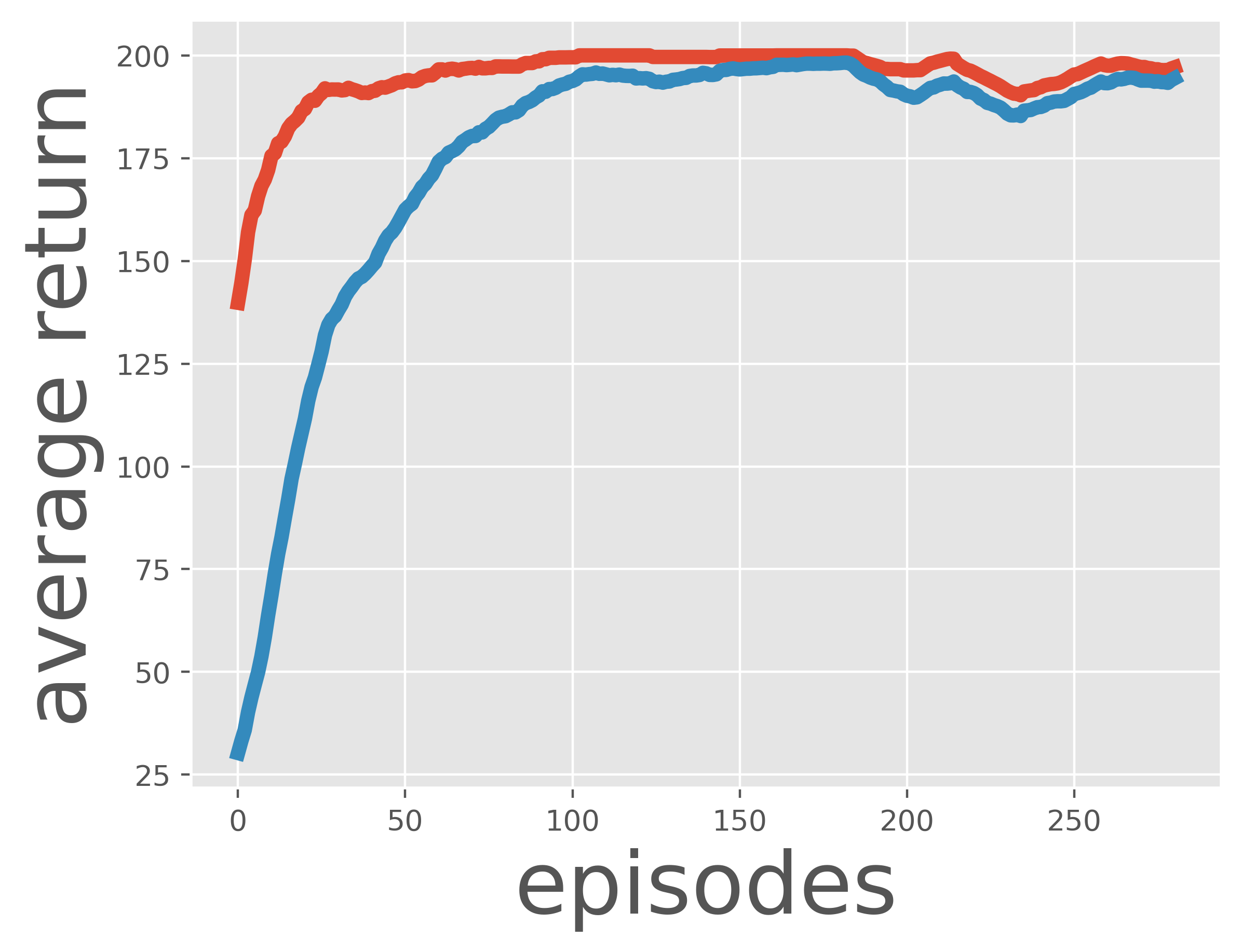}
}
\caption{Average overall reward of $\textsc{DQInit}(\widetilde\pi,\widetilde{\mathcal{L}},\mathcal{L}_{KL})$ vs  $\theta$-dependent average return ($\mathcal{R}_{\text{avg}}(\theta)$) in different environments}\label{fig:theta}
\end{figure}
\subsection{Empirical convergence proof}

We introduce a novel metric, the $\theta$-dependence reward, designed to measure the alignment between the overall return $\mathcal{R}_{\text{avg}}$ and the agent’s reliance on its value estimates. It is defined as:
\begin{equation*}
\mathcal{R}_{\text{avg}}(\theta) = \left( \frac{\sum_{(s,a) \sim \tau} K(s,a)}{\text{episode steps}} \right) \times \mathcal{R}_{\text{avg}}
\end{equation*}

This metric quantifies the degree to which high returns coincide with the agent acting on its internal knowledge, as expressed by the episodic average knownness, $\frac{\sum_{(s,a) \sim \tau} K(s,a)}{\text{episode steps}}$ which ranges from 0 (all visited states are unknown) to 1 (all visited states are known). A value of $\mathcal{R}_{\text{avg}}(\theta)$ close to $\mathcal{R}_{\text{avg}}$ suggests that the agent is both achieving strong performance increasingly based on its learned function, rather than on the transferred value.
We report the evolution of this metric alongside the total reward in Figure~\ref{fig:theta}, using \textsc{LogQInit} for MountainCar and Acrobot, and $\mathsf{UCOI}$ for CartPole. The results show that dependence on internal model parameters grows over time, eventually matching the overall performance, indicating convergence of the learning process.
\subsection{Robustness in Sparse Reward Settings}
A notable observation is that in both MountainCar and Acrobot, the target tasks used for transfer featured extremely sparse reward structures—binary rewards of 0 for all transitions and 1 only upon reaching the goal. Despite this, $\textsc{DQInit}$ was consistently able to converge and solve the tasks effectively. This highlights the robustness of the initialization strategy and its ability to guide early exploration, even when reward signals are delayed or rare. We emphasize that no reward shaping was used, and the agent learned purely from sparse binary returns.

\subsection{The choice of $K$ parameters:} The knownness function plays a central role in modulating the influence of the initialization function throughout learning. Its rate of increase is governed by the parameters $m$ and $p$, which must be carefully selected to balance two competing objectives: (1) allowing the agent sufficient time to explore and learn about each state-action pair before the initialization influence fades, and (2) ensuring that, once the knownness converges toward 1, the agent can produce a reliable and beneficial policy from that state. If knownness increases too quickly, initialization influence may diminish before the agent has adequately learned a good policy. On the other hand, if it increases too slowly, the initialization persists longer than necessary, potentially hindering the learning of a task-specific value function.

To determine appropriate values for $m$ and $p$, we conducted an empirical analysis across a range of candidate values using multiple evaluation metrics. The results are summarized in Table~\ref{tab:mp}, with three key metrics reported: average return ($R_{\text{avg}}$), average $\theta$-dependence reward ($R_{\text{avg}}(\theta)$), and knownness percentage ($K\%$). Metrics annotated with [-N] indicate averages computed over the final $N$ episodes, which helps assess whether high knownness levels coincide with effective policy performance or indicate suboptimal convergence.

Our goal is to strike a balance between sufficient exploration and the timely fading of initialization influence. For this reason, we occasionally favor parameter settings that yield slightly lower returns in exchange for more stable knownness convergence and interpretable learning dynamics.

Our experimental results across MountainCar, CartPole, and Acrobot demonstrate the nuanced relationship between knownness parameters $(p, m)$ and learning performance, as measured through both standard average reward (solid bars) and $\theta$-dependence reward (hatched bars).

\begin{table}[ht]
\centering
\caption{Experiments configuration details}\label{tab:expdetails}
\resizebox{\textwidth}{!}{\begin{tabular}{|c|c|c|c|}
\hline
Parameters                  & Mountain car                               & Acrobot                        & Cartpole                                              \\ \hline
Network                     & \multicolumn{3}{c|}{32 x 3}                                                                                                          \\\hline
Batch                       & 64                                         & 128                            & 64                                                    \\\hline
Memory capacity             & 300.000                                    & 10.000                         & 10.000                                                \\\hline
Learning rate               & 0.001                                      & 0.0001                         & 0.001                                                 \\\hline
No. tasks b/transfer    & \multicolumn{3}{c|}{30}                                                                                                              \\\hline
Episodes b/transfer          & 2000                                       & 2000                           & 1000                                                  \\\hline
No. tasks a/tansfer         & \multicolumn{3}{c|}{30}                                                                                                              \\\hline
Episodes a/transfer        & 1000                                       & 600                            & 100                                                   \\\hline
Reward b/transfer           & shaped                                     & If s=g -1, 0   otherwise       & \multirow{3}{*}{0 if state is penalizing 1 otherwise} \\\hline
Reward of ($Q^{\#}$)b/transfer    & \multicolumn{2}{c|}{\multirow{2}{*}{Binary positive r(g)= 1, 0 otherwise}} &                                                       \\
Reward a/transfer                & \multicolumn{2}{c|}{}                                                        &                                                       \\\hline
discrete state $\phi$                         & Interval $\hat{S}=30\times30$        &      $\hat{S}=20\times20\times 20 \times 20$  & $\hat{S}=20\times20\times 20 \times 20$ \\ \hline
Maxstep                     & 300                                        & 500                            & 200                                                   \\\hline
m,p & 100, 10				& 50, 10			&20, 1\\\hline
Max ep, decay min   b/trans & \multicolumn{3}{c|}{(1, 0.999/step, 0.01)}
\\\hline
\end{tabular}}
\end{table}

\begin{figure*}[ht!]
\centering 
\begin{tabular}{m{0.1em}*{4}{>{\centering\arraybackslash}m{10em}}}

&LogQinit & UCOI & MaxQinit\\
\rotatebox{90}{Mountain Car} &\includegraphics[width=1 \linewidth]{mc_logmean}
 &\includegraphics[width=1\linewidth]{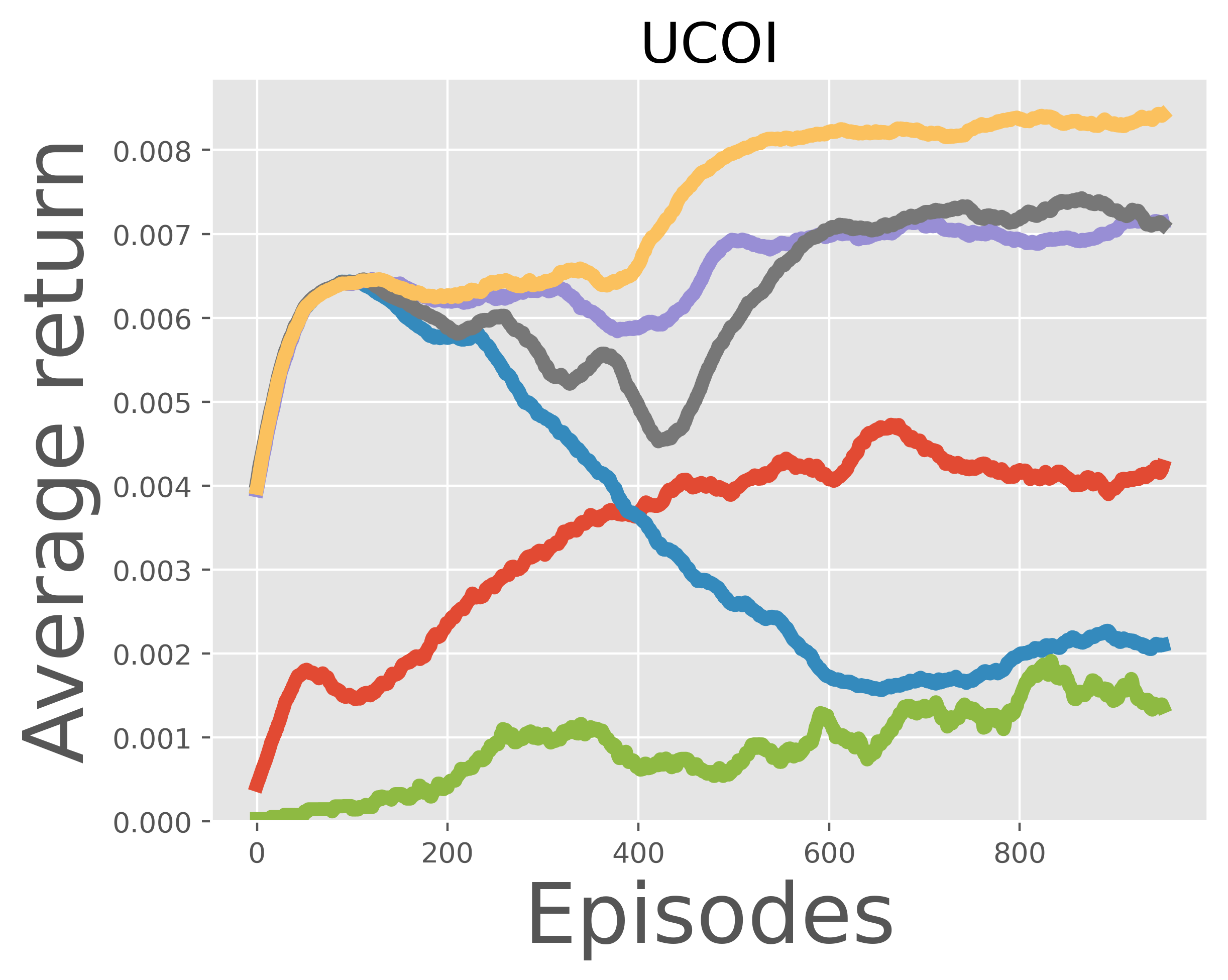}&\includegraphics[width=1 \linewidth]{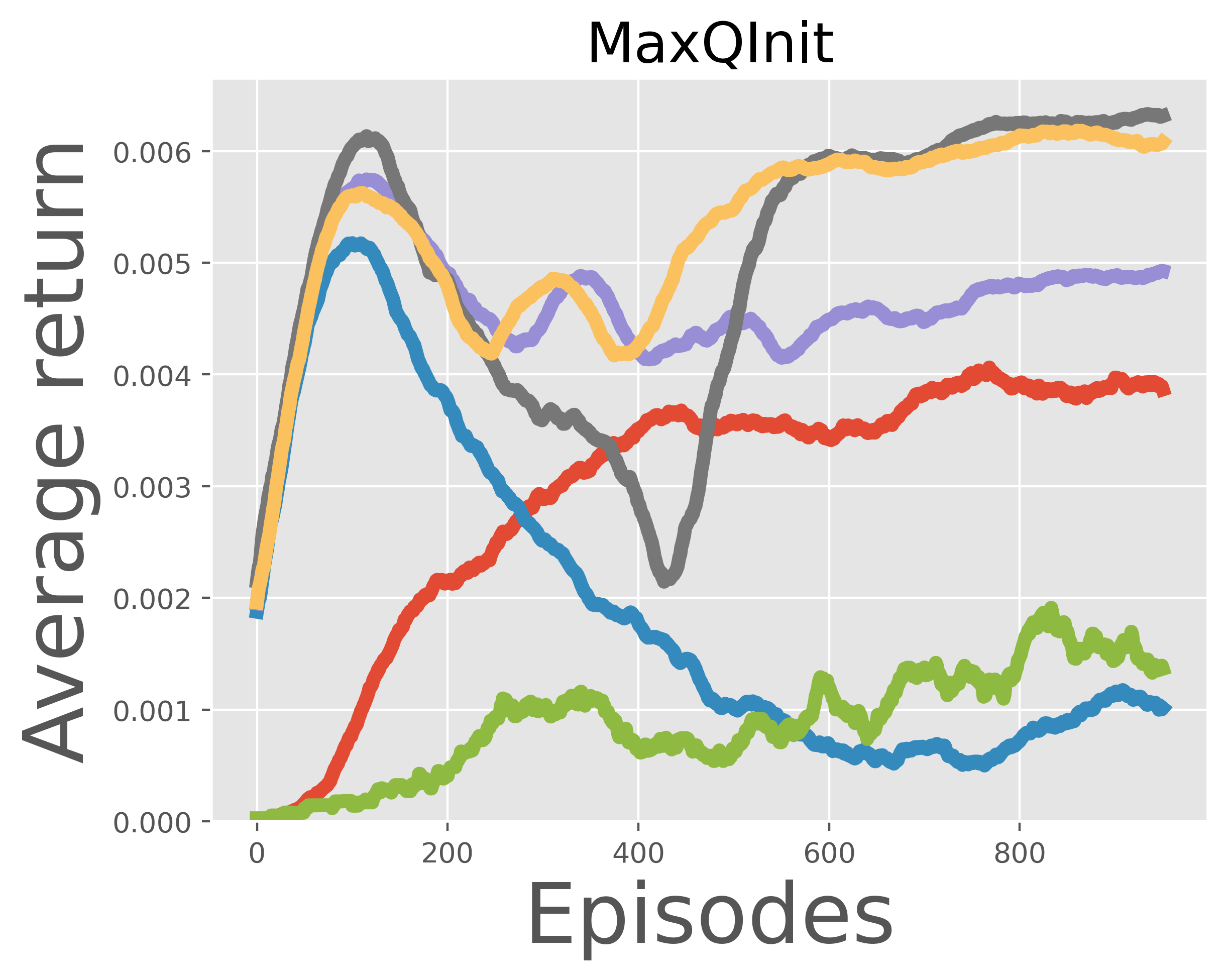} \\
\rotatebox{90}{Acrobot} &\includegraphics[width=1 \linewidth]{ac_logmean}
 &\includegraphics[width=1\linewidth]{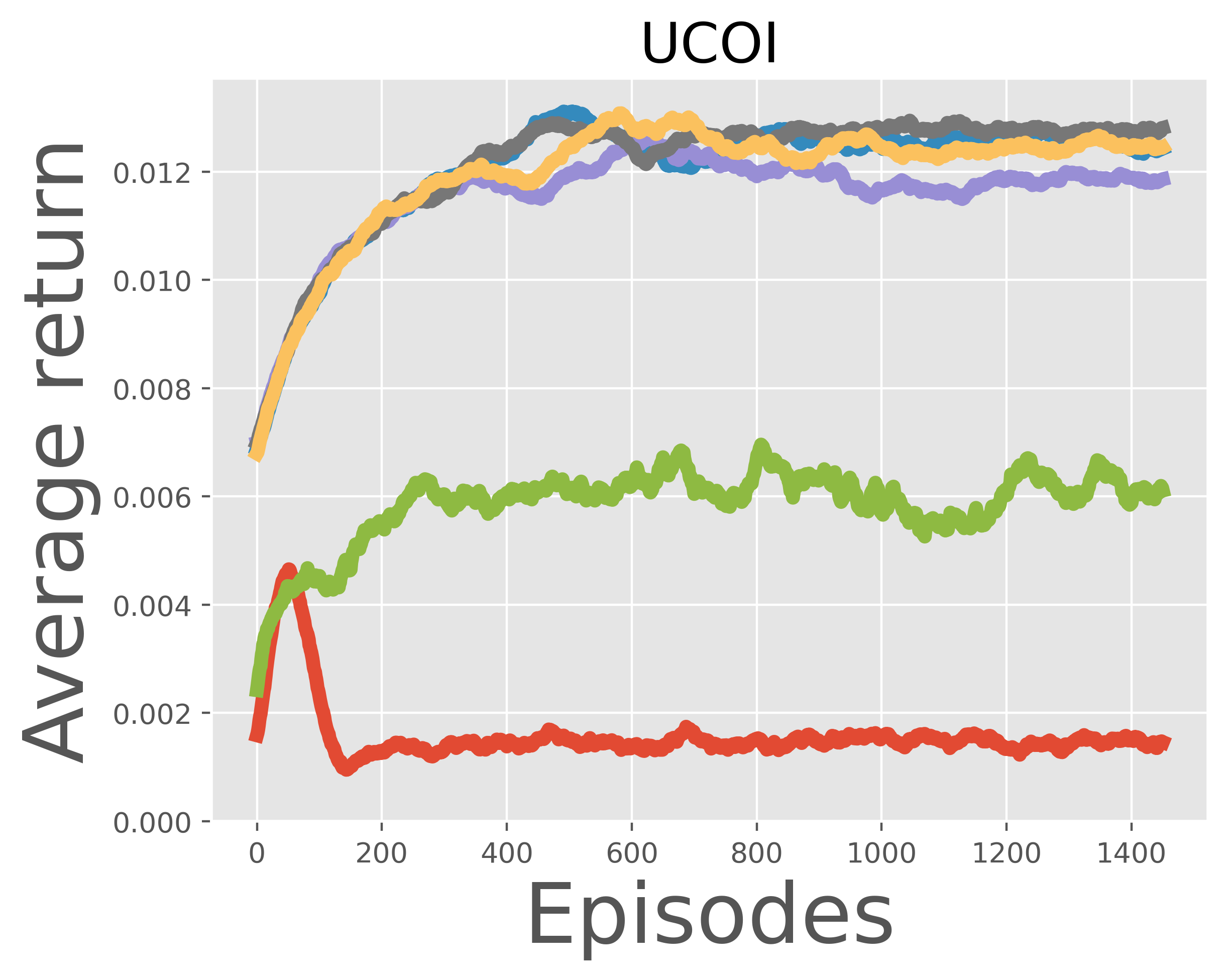}&\includegraphics[width=1 \linewidth]{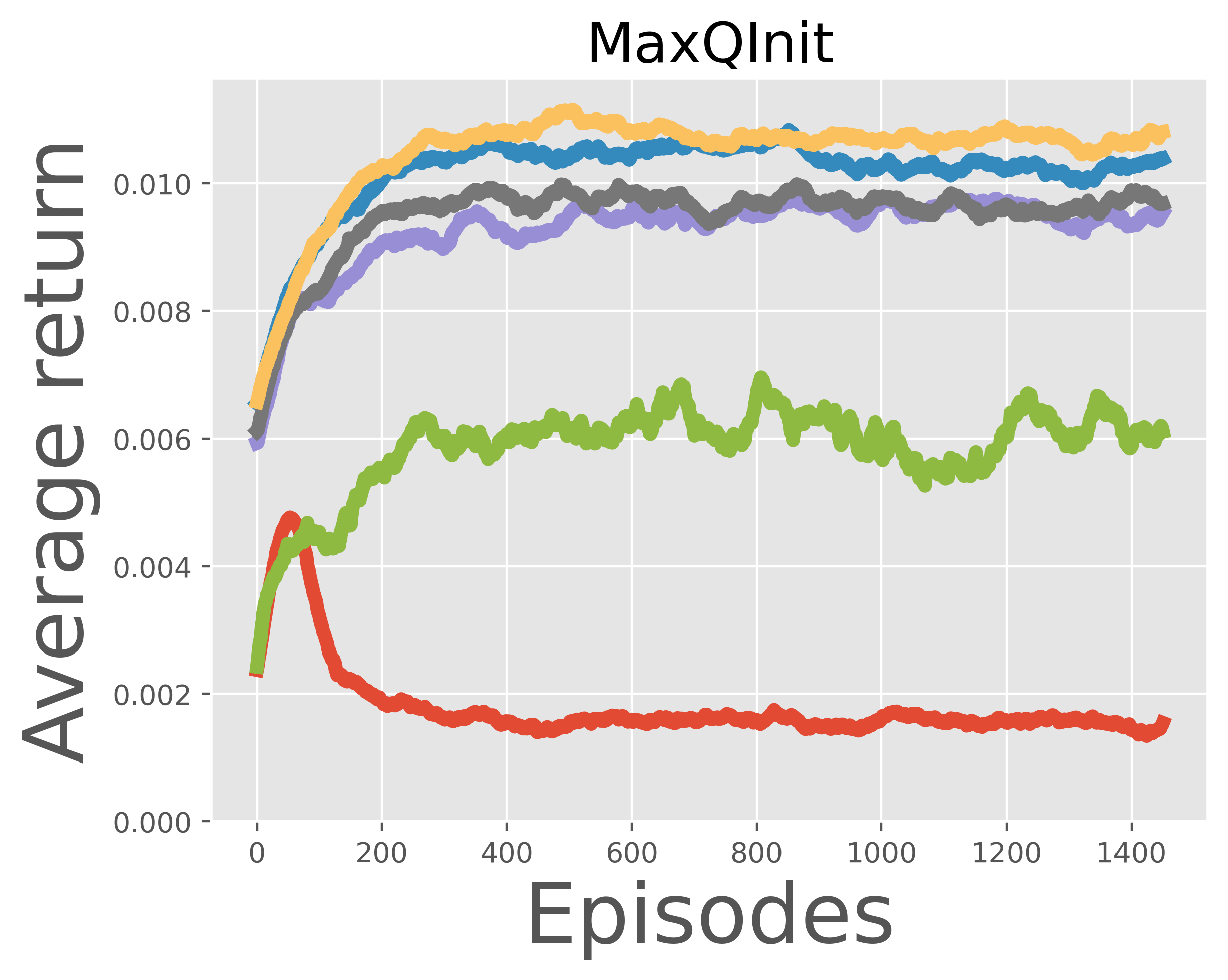}\\
\rotatebox{90}{Cartpole}&\includegraphics[width=1 \linewidth]{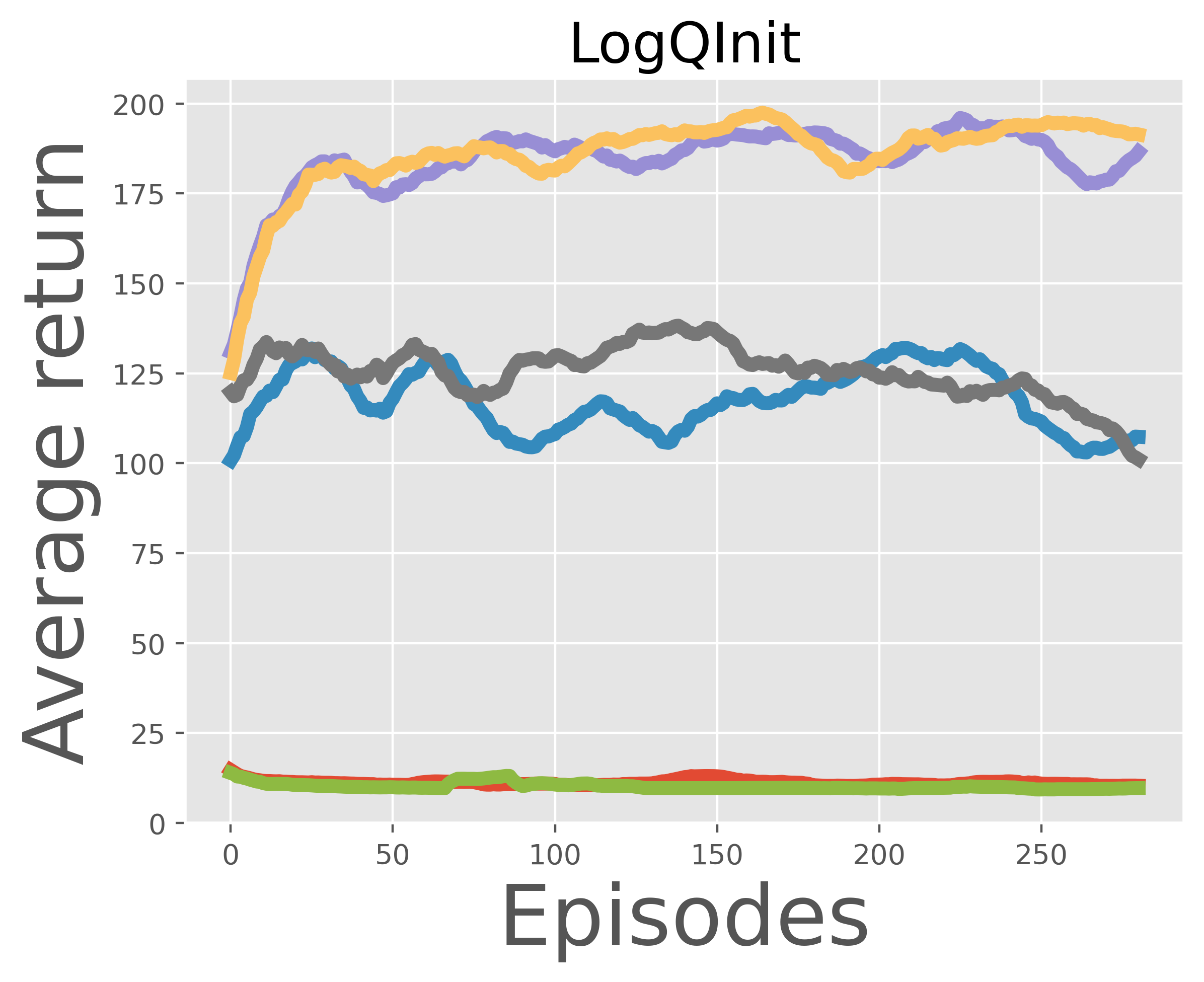}
 &\includegraphics[width=1\linewidth]{cp_mean}&\includegraphics[width=1 \linewidth]{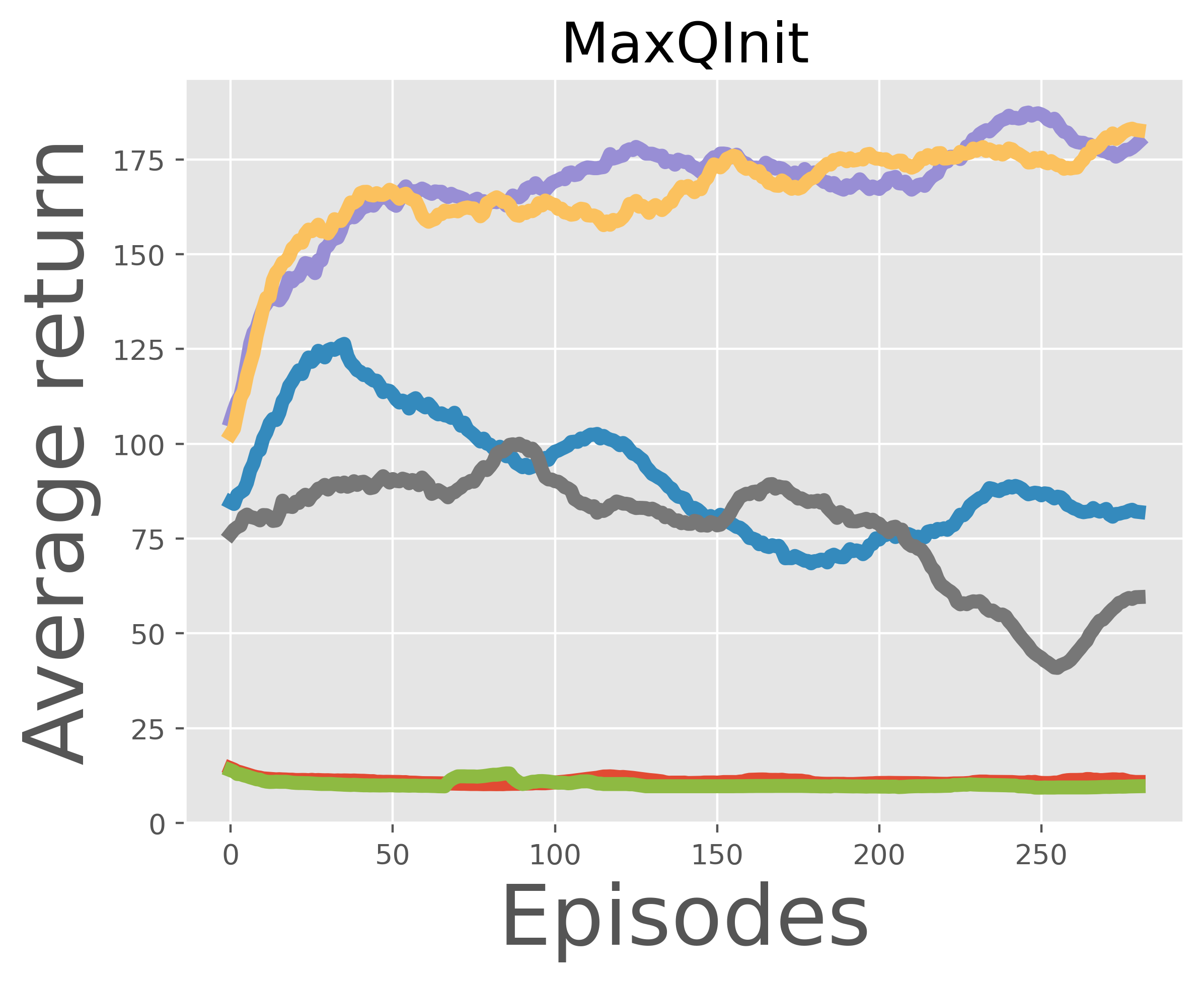}
\end{tabular}
\includegraphics[width=.8\linewidth]{init}
\caption{
Performance of $\textsc{DQInit}$ across initialization strategies (MaxQInit, UCOI, LOGQInit) and usage modes ($\tilde{\pi}$, $\tilde{L}$, $L_{\text{KL}}$). LOGQInit performs best in MountainCar, UCOI in CartPole. Combining all modes yields the most stable performance across environments.}\label{fig:init}	
\end{figure*}

\begin{table*}
\centering
\caption{average reward and model dependence average reward of different values of parameter m and p. The results were averaged over 10 tasks of each environment}\label{tab:mp}
\resizebox{\textwidth}{!}{\begin{tabular}{|l|l|l|l|l|l|l|l|l|l|l|}
                                                              \cline{3-11}\multicolumn{2}{l}{~ }   & \multicolumn{3}{|c}{~ ~ Acrobot} & \multicolumn{3}{|c}{~ ~ Cartpole} & \multicolumn{3}{|c|}{~ ~ Mountain car}  \\ \hline
   p$\downarrow$ & m$\rightarrow$     & 20     & 50     & 100           & 20     & 50     & 100            & 20      & 50     & 100                \\\hline
\multirow{4}{*}{4}                                                             & $\mathcal{R}_{avg}$  & 0.0098 & 0.0111 & 0.0115        & 171.61 & 177.19 & 180.73         & 0.00721 & 0.0073 & 0.0076             \\
                                                                               & $\mathcal{R}_{avg}[-100]$ & 0.0099 & 0.0116 & 0.0118        & 190.47 & 185.76 & 192.91         & 0.008   & 0.009  & 0.009              \\
                                                                               & $\mathcal{R}_{avg}(\theta)$    & 0.0066 & 0.0063 & 0.0058        & 75.81  & 44.65  & 25.44          & 0.0068  & 0.0064 & 0.0061             \\
                                                                               & $\mathcal{R}_{avg}(\theta)[-100]$       & 0.0080 & 0.0089 & 0.0080        & 140.00 & 79.68  & 59.21          & 0.0080  & 0.0089 & 0.0088             \\\hline
\multirow{4}{*}{10}                                                            & $\mathcal{R}_{avg}$       & 0.0094 & 0.0115 & 0.0015        & 174.11 & 174.87 & 181.25         & 0.0071  & 0.0071 & 0.0077             \\
                                                                               & $\mathcal{R}_{avg}[-100]$    & 0.0098 & 0.0117 & 0.0017        & 190.92 & 183.63 & 192.81         & 0.0083  & 0.0087 & 0.0089             \\
                                                                               & $\mathcal{R}_{avg}(\theta)$     & 0.0061 & 0.0062 & 0.0056        & 81.67  & 26.67  & 19.16          & 0.0067  & 0.0062 & 0.0061             \\
                                                                               & $\mathcal{R}_{avg}(\theta)[-100]$   & 0.0078 & 0.0090 & 0.0079        & 151.11 & 55.29  & 37.34          & 0.0083  & 0.0086 & 0.0087             \\\hline
\multirow{4}{*}{\begin{tabular}[c]{@{}l@{}}20 (1 for\\ cartpole)\end{tabular}} & $\mathcal{R}$    & 0.0097 & 0.0111 & 0.0115        & 178.94 & 183.6  & 184.29         & 0.0071  & 0.0074 & 0.0071             \\
                                                                               & $\mathcal{R}[-100]$      & 0.0097 & 0.0115 & 0.0119        & 197.12 & 195.6  & 196.98         & 0.0078  & 0.0089 & 0.079              \\
                                                                               & $\mathcal{R}_{avg}(\theta)$   & 0.0062 & 0.0059 & 0.0055        & 113.99 & 90.56  & 59.94          & 0.0066  & 0.0063 & 0.0055             \\
                                                                               & $\mathcal{R}_{avg}(\theta)[-100]$    & 0.0077 & 0.0085 & 0.0080        & 173.59 & 146.82 & 101.78         & 0.0078  & 0.0088 & 0.0077\\
\hline   
\end{tabular}}
\end{table*}

\section{Conclusion}

In this work, we addressed the limitations of VFI strategies in transfer reinforcement learning, which have so far been confined to tabular settings. To extend these benefits to DRL, we proposed \textsc{DQInit}, a novel approach that integrates transferred knowledge into the DRL value function via a knownness-based mechanism over clustered state-action pairs. Our method supports multiple usage modes, including policy-guided action selection, value alignment, and distillation.

$\textsc{DQInit}$ effectively combines the jumpstart benefits of policy guidance with the long-term stability of distillation, outperforming standard baselines and reproducing the performance trends observed in tabular VFI methods. Moreover, by using tabular value functions as compact knowledge sources, $\textsc{DQInit}$ mitigates the instability commonly seen in teacher network outputs, improving both reliability and storage efficiency. Notably, $\textsc{DQInit}$ was also able to solve sparse binary reward, demonstrating strong exploratory guidance even in environments with delayed or minimal feedback.

\section*{Limitations}

First, we note that our experiments were designed primarily for analytical purposes, serving as an initial investigation into the adaptability of value function initialization in deep reinforcement learning. For this reason, we evaluated $\textsc{DQInit}$ only within the DQN framework and on classical control tasks. However, $\textsc{DQInit}$ is modular and readily adaptable to other DRL methods, and a natural extension would be to evaluate it in actor-critic and policy optimization algorithms.

A second limitation lies in the discretization of the state-action space. While we employed simple clustering methods in this work, more rigorous state abstraction techniques should be considered in future work to reduce the risk of over-generalization and to ensure the accuracy of the value function used for transfer.

Lastly, the knownness function plays a central role in $\textsc{DQInit}$’s integration mechanism. Further investigation is needed to understand how to best define and tune this parameter to balance the influence of transferred and learned knowledge effectively.

\bibliographystyle{plain}
\bibliography{aaai2026}

\end{document}